%% file: main.tex
\documentclass[10pt,letterpaper]{article}
\input{preamble}
\usepackage{multicol}
\usepackage{stfloats}
\usepackage{tikz}
\usepackage{colortbl}
\usetikzlibrary{arrows.meta}

\title{PySWBench: A Native CAD Generation Benchmark with Agentic Evaluation}
\date{}
\newcommand{\PySWAffiliations}[1][]{\textsuperscript{1#1}}
\author{%
  Hui Deng\PySWAffiliations[\textasteriskcentered], Sicheng He\PySWAffiliations[\textasteriskcentered],
  Jintao Li\PySWAffiliations, Chen Zhang\PySWAffiliations,\\
  Peiji Li\PySWAffiliations, Shui Yu\PySWAffiliations[\textdagger], Yun Li\PySWAffiliations[,2\textdagger]\\
  \textsuperscript{1}University of Electronic Science and Technology of China\\
  \textsuperscript{2}i4AI Ltd and i4AI Centre
}
\hypersetup{
  pdftitle={PySWBench: A Native CAD Generation Benchmark with Agentic Evaluation},
  pdfauthor={Hui Deng, Sicheng He, Jintao Li, Chen Zhang, Peiji Li, Shui Yu, Yun Li}
}

\begin{document}
\twocolumn
\setcounter{dbltopnumber}{2}
\setcounter{topnumber}{1}
\renewcommand{\dbltopfraction}{0.9}
\renewcommand{\topfraction}{0.9}
\renewcommand{\textfraction}{0.08}
\setlength{\textfloatsep}{12pt plus 2pt minus 2pt}
\setlength{\dbltextfloatsep}{12pt plus 2pt minus 2pt}
\flushbottom
\clubpenalty=1000
\widowpenalty=1000
\maketitle
\begingroup
\renewcommand{\thefootnote}{\fnsymbol{footnote}}
\footnotetext[1]{Equal contribution.}
\footnotetext[2]{Co-corresponding authors.\\
  \hspace*{1.5em}\textit{E-mail addresses:} \href{mailto:Shui.Yu@i4ai.org}{Shui.Yu@i4ai.org},
  \href{mailto:Yun.Li@ieee.org}{Yun.Li@ieee.org}.}
\endgroup
\input{sec/0_abstract}
\input{sec/1_introduction}
\input{sec/2_related_work}
\input{sec/3_benchmark}
\input{sec/4_experiments}
\input{sec/5_conclusion}

\input{sec/appendices}

\end{document}

%% file: preamble.tex
\usepackage[letterpaper,top=1in,textheight=8.875in,textwidth=6.875in,nohead]{geometry}
\usepackage{times}
\usepackage[dvipsnames]{xcolor}
\usepackage{titling}
\usepackage{titlesec}
\usepackage[font=small,labelsep=period]{caption}

\pretitle{\begin{center}\Large\bfseries}
\posttitle{\par\end{center}\vspace{4pt}}
\predate{}
\postdate{}

\titleformat{\section}{\large\bfseries}{\thesection.}{0.5em}{}
\titleformat{\subsection}{\normalsize\bfseries}{\thesubsection.}{0.5em}{}
\titleformat{\subsubsection}{\normalsize\bfseries}{\thesubsubsection.}{0.5em}{}
\titlespacing*{\section}{0pt}{10pt plus 2pt minus 2pt}{7pt}
\titlespacing*{\subsection}{0pt}{8pt plus 2pt minus 2pt}{5pt}
\titlespacing*{\subsubsection}{0pt}{6pt plus 2pt minus 2pt}{3pt}

\ifdefined\XeTeXversion\usepackage[T1]{fontenc}\fi

\input{math_commands.tex}

\usepackage{float}

\usepackage[hyphens]{url}
\usepackage{graphicx}
\usepackage{flafter}
\usepackage{placeins}
\usepackage{booktabs,tabularx,amssymb,multirow}

\usepackage{amsthm}
\usepackage{listings}
\lstdefinestyle{pyswexample}{
    language=Python,
    basicstyle=\small\ttfamily,
    keywordstyle=\bfseries,
    commentstyle=\itshape,
    numbers=left,
    numberstyle=\scriptsize,
    numbersep=8pt,
    xleftmargin=2em,
    framexleftmargin=1.6em,
    frame=tb,
    framerule=0.4pt,
    framesep=5pt,
    columns=fullflexible,
    keepspaces=true,
    showstringspaces=false,
    breaklines=true,
    captionpos=b,
    aboveskip=8pt,
    belowskip=8pt
}

\usepackage[numbers,sort&compress]{natbib}
\usepackage{microtype}
\definecolor{paperlinkblue}{rgb}{0.21,0.49,0.74}
\usepackage[pagebackref,breaklinks,colorlinks,allcolors=paperlinkblue]{hyperref}

%% file: math_commands.tex
\usepackage{amsmath,amsfonts,bm}

\def\eqref#1{equation~\ref{#1}}

\def\1{\bm{1}}

\DeclareMathAlphabet{\mathsfit}{\encodingdefault}{\sfdefault}{m}{sl}
\SetMathAlphabet{\mathsfit}{bold}{\encodingdefault}{\sfdefault}{bx}{n}



%% file: sec/0_abstract.tex
\begin{abstract}
\itshape
\setlength{\baselineskip}{11pt}
\noindent
Large language models (LLMs) and agents can now generate CAD models from text
descriptions and engineering drawings, making the assessment of their modeling
capabilities increasingly important. Geometric similarity captures overall
shape differences but reveals little about whether specific design requirements
are met or where modeling errors arise. We introduce PySWBench, a native CAD
benchmark for common mechanical parts, covering text-to-CAD,
drawing-to-CAD, and within-family modification. Our data pipeline recovers
modeling programs from SolidWorks parts and verifies the rebuilt models, while
PySW provides tools for modeling and inspection. An evaluation agent combines
programs, execution logs, and native models to check design requirements,
assess six process dimensions, and trace error sources and their downstream
effects. We evaluate seven LLMs on 150 parts and 40 modification tasks,
alongside two models paired with two agent harnesses. All models achieve lower
requirement satisfaction on drawings than on text, with requirement
interpretation and dimension setting as shared weaknesses. Process analysis
links some dimensional and contour errors to misunderstood requirements and
shows how planning errors affect subsequent feature construction.
Feedback-based revisions improve satisfaction, but gains depend on the
model--harness pairing and can come with new errors. Against human annotations,
our evaluator achieves accuracies of 99.35\% and 93.67\% on two test sets, with
the lowest miss and false-alarm rates among the evaluated methods.
\par
\end{abstract}

%% file: sec/1_introduction.tex
\section{Introduction}
\label{sec:introduction}

Computer-aided design (CAD) is essential to product design and manufacturing.
Engineers spend considerable time creating and revising CAD
models~\citep{willis2021fusion}. Large language models (LLMs) and agents can now
generate these models from text descriptions and engineering
drawings~\citep{xie2025text,wang20252d,mallis2025cad}.
Their outputs are often evaluated by geometric similarity to a reference
model~\citep{wu2021deepcad,doris2026cad}. However, a close overall match can still
hide missing features or dimensional errors that violate design requirements.
Evaluation must therefore assess how well generated CAD models meet the
stated design requirements.

Existing CAD benchmarks assess design requirements, modeling operations, and
editing results through code analysis, rendered views, and executable geometric
tests~\citep{zhang2026benchcad,dong2025histcad,du2024blenderllm,mallis2026text}.
These checks, however, do not provide equivalent evidence. Parameters and
operations in code may fail to produce the intended geometry, rendered views
cannot replace precise measurements, and final geometry alone cannot reveal how
a solid was built~\citep{singh2026cadengbench,mallis2026text,dong2026cadworld}.
The underlying limitation is that native sketches, driving dimensions, and
feature relationships are often not preserved or examined, leaving construction
to be inferred from code or shape. Recent work also uses predefined rules to inspect native
documents~\citep{dong2026cadworld}. However, the same requirement may be met
through different constructions, which may call for different objects and
measurements to verify it~\citep{mallis2026text,jones2023b}. Evaluation should
therefore combine native construction records with actual geometry, keeping
acceptance criteria consistent while adapting verification to each candidate's
implementation.

Evaluation should also identify weaknesses in the modeling process. Existing
benchmarks analyze operation use, parameter response, preservation, and failure
types~\citep{zhang2026benchcad,singh2026cadengbench,dong2026cadworld}. Operation
recall and individual failure labels, however, do not fully capture concurrent
errors across modeling stages or their relationships. Final geometric deviations
may originate in earlier requirement interpretation, planning, or sketch
construction. Combining code, execution records, and native CAD models to assess
these stages and trace error propagation helps locate error sources
and steps requiring correction.

To this end, we introduce PySWBench, a native CAD benchmark for generating
common mechanical parts from text descriptions (Text2CAD) and engineering drawings
(Drawing2CAD) and modifying parts within design families. PySW wraps the SolidWorks application programming interface
(API) to expose construction records and precise
measurements. An evaluation agent combines construction records
and actual geometry to verify requirements and assess six modeling
process dimensions. It traces error sources and downstream effects using evidence.

We evaluate seven LLMs on text and drawing inputs for 150 common mechanical
parts and on 40 modification tasks. All seven achieve lower requirement satisfaction on drawings, where their
two weakest areas are capturing all design requirements and assigning the
correct dimensions to the relevant sketch entities. Process analysis further links some dimensional and contour errors to misinterpreted
requirements, while planning errors can affect subsequent feature construction:
in both generation settings, over half of the outputs with planning failures
record propagation to feature construction. Feedback-based revisions improve satisfaction, but gains depend on the
model--harness pairing and can come with new errors. Against human annotations, our evaluator achieves 99.35\% and 93.67\%
accuracy on the two test sets, with
the lowest miss and false-alarm rates among the evaluated methods.
This work makes three main contributions:

\begin{itemize}
    \setlength{\itemsep}{3pt}
    \setlength{\parskip}{0pt}
    \item \textbf{A native CAD benchmark and data pipeline.} We introduce
    PySWBench for text-to-CAD, drawing-to-CAD, and within-family modification
    of common mechanical parts. A conversion pipeline recovers
    and verifies modeling programs from native SolidWorks parts, providing
    paired programs, models, and views, alongside PySW tools for modeling
    and inspection.
    \item \textbf{Process metrics and error propagation.} We assess six process dimensions
    across modeling strategy, 2D geometry, and 3D construction to identify modeling
    weaknesses and quantify evidence-supported error propagation across stages.
    \item \textbf{Agentic evaluation with native evidence.} Our evaluation agent
    combines programs, execution logs, and native CAD models to verify
    requirements against a shared task checklist and trace error sources.
    Comparisons with human annotations validate its reliability.
\end{itemize}

%% file: sec/2_related_work.tex
\begin{figure*}[!t]
    \centering
    \includegraphics[width=\textwidth,trim=49bp 7.5bp 31.4bp 14.7bp,clip]{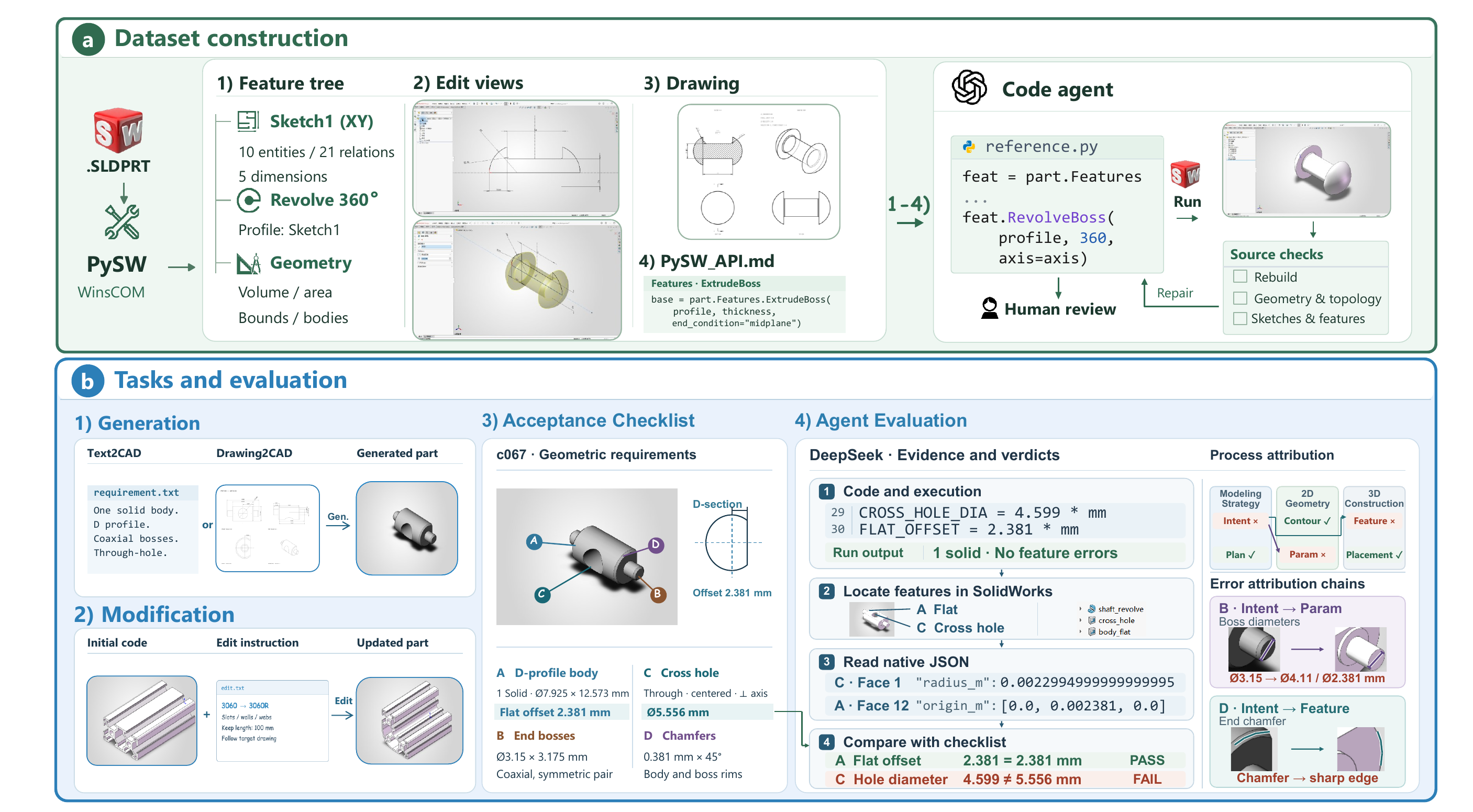}
    \caption{PySWBench construction and evaluation.}
    \label{fig:architecture}
\end{figure*}

\section{Related Work}
\label{sec:related-work}

\input{tables/table1}

\paragraph{CAD representations.}
The CAD representation determines what an evaluator can inspect. Point-cloud
and mesh generators produce geometry without explicit modeling
histories~\citep{yang2019pointflow,nash2020polygen,luo2021diffusion,siddiqui2024meshgpt}.
Sequence-based methods retain sketch-and-extrude operations but typically use
a limited operation vocabulary~\citep{wu2021deepcad,willis2021fusion,xu2022skexgen,xu2023hierarchical,khan2024cad,khan2024text2cad,chen2025cadcrafter}.
CadQuery-based methods represent CAD models as executable parameterized
programs~\citep{xie2025text,guan2026cad,rukhovich2025cad}.
Other methods directly generate B-rep geometry and
topology~\citep{jayaraman2022solidgen,xu2024brepgen}.
STEP preserves geometry but generally drops sketch constraints, driving dimensions,
and feature dependencies~\citep{pratt2005towards,dong2026cadworld}. Boundary geometry
alone cannot uniquely recover these records. HistCAD explicitly preserves
constraint-aware parametric construction histories~\citep{dong2025histcad}.

\paragraph{CAD evaluation.}
CADBench and IterCAD-Bench primarily evaluate shape validity and geometric
similarity~\citep{doris2026cadbench,hu2026itercad}. Text2CAD-Bench also uses visual
scoring to check design requirements~\citep{wang2605text2cad}.
BlenderLLM checks attributes and spatial relations from scripts and
views~\citep{du2024blenderllm}; script parameters still require verification against
executed geometry, and views cannot replace precise measurements. CADTests
refines executable B-rep tests through mutation analysis~\citep{mallis2026text}.
MUSE uses component matching and interface measurements to check connectivity,
interface types, and clearances, and a visual judge to assess functional
consistency~\citep{dong2026muse}. RealCADBench evaluates exported models through
execution checks, geometric similarity, and visual-semantic
judgments~\citep{li2026realcadbench}. CADWorld inspects native properties,
constraints, and construction structure using task-specific programs that
predefine object selection and property checks~\citep{dong2026cadworld}.

Process analysis further localizes modeling errors. BenchCAD analyzes capability
weaknesses through operation recall, edit-failure categories, and paired question
answering~\citep{zhang2026benchcad}. HistCAD studies editability through constraint
restoration experiments and failure-stage analysis~\citep{dong2025histcad}.
CADEngBench tests parameter response, functional edits, and
preservation~\citep{singh2026cadengbench}. CADWorld assigns each failed episode
its first applicable failure category~\citep{dong2026cadworld}. These analyses
identify weaknesses or error sources, but their evaluation protocols do not
explicitly record error propagation across modeling stages. PySWBench combines
evidence from programs, execution logs, and native models to connect requirement
verification with error propagation analysis. Table~\ref{tab:intro-motivation}
summarizes the differences between these evaluation protocols.

%% file: tables/table1.tex
\begin{table}[!t]
    \centering
    \caption{Comparison of CAD evaluation protocols.}
    \label{tab:intro-motivation}
    \begingroup
    \fontsize{8}{9.5}\selectfont
    \setlength{\tabcolsep}{2pt}
    \renewcommand{\arraystretch}{1.12}
    \begin{tabular*}{\columnwidth}{@{\extracolsep{\fill}}lcccc@{}}
        \toprule
        Method & Requirements & Native & Process & Propagation \\
        \midrule
        CADBench \citep{doris2026cadbench} & -- & -- & -- & -- \\
        Text2CAD-Bench \citep{wang2605text2cad} & $\checkmark$ & -- & -- & -- \\
        IterCAD-Bench \citep{hu2026itercad} & -- & -- & -- & -- \\
        BenchCAD \citep{zhang2026benchcad} & $\checkmark$ & -- & $\checkmark$ & -- \\
        BlenderLLM \citep{du2024blenderllm} & $\checkmark$ & -- & -- & -- \\
        CADTests \citep{mallis2026text} & $\checkmark$ & -- & -- & -- \\
        HistCAD \citep{dong2025histcad} & $\checkmark$ & $\checkmark$ & $\checkmark$ & -- \\
        CADEngBench \citep{singh2026cadengbench} & $\checkmark$ & $\checkmark$ & $\checkmark$ & -- \\
        MUSE \citep{dong2026muse} & $\checkmark$ & -- & $\checkmark$ & -- \\
        RealCADBench \citep{li2026realcadbench} & $\checkmark$ & -- & -- & -- \\
        CADWorld \citep{dong2026cadworld} & $\checkmark$ & $\checkmark$ & $\checkmark$ & -- \\
        \midrule
        \textbf{PySWBench (ours)} & $\checkmark$ & $\checkmark$ & $\checkmark$ & $\checkmark$ \\
        \bottomrule
    \end{tabular*}
    \par\smallskip
    \begin{minipage}{\columnwidth}
    \raggedright
    Requirements: checks whether design requirements are met;
    Native: inspects construction records in native CAD documents;
    Process: diagnoses modeling operations, stages, or structured failure types;
    Propagation: explicitly records evidence-supported error links across
    modeling stages.
    \end{minipage}
    \endgroup
\end{table}

%% file: sec/3_benchmark.tex
\section{PySWBench}
\label{sec:benchmark}
\label{sec:evaluation}

\subsection{Dataset Construction}
\label{sec:data-construction}
\label{sec:native-environment}
\label{sec:construction}

PySW extracts sketch dimensions, feature parameters, and
dependencies from native SolidWorks part files (\texttt{.SLDPRT}), together with
engineering drawings and edit-mode views. A recovery agent
uses these records and PySW API documentation to reconstruct
each part as a PySW program.
We execute the programs in SolidWorks and compare the rebuilt
parts with their sources, checking rebuild status, body count,
bounding box, volume, surface area, topology, and construction
records to guide further revision. Figure~\ref{fig:architecture}(a) summarizes
this reconstruction and verification workflow.

\subsection{Task Definition}
\label{sec:task-definition}

\paragraph{Generation.}
We build text and drawing generation tasks for the same
150 common mechanical parts, each serving a practical engineering function. An agent drafts
text specifications from feature trees, model views, and sketch
views in edit mode and adds dimensions to native drawings,
after which we manually verify the specifications and annotations.
LLMs receive text or a drawing with PySW API documentation
and generate a complete PySW program. Figure~\ref{fig:architecture}(b)
shows examples of generation and modification.
The generation tasks above evaluate models' end-to-end modeling ability.
For agent generation, we select 40 more challenging tasks from the
150 Text2CAD tasks to stress-test agents' ability to revise their programs
using tool feedback. Each task allows at most three build attempts,
including failed attempts.

\paragraph{Modification.}
We construct 40 modification tasks by pairing initial and
target parts within the same design families. LLMs receive
an initial PySW program, part views, edit instructions, and
API documentation, and generate a complete revised program.

\subsection{Evaluation Framework}
\label{sec:evaluation-framework}

\input{tables/table2}

We assess how well CAD models meet design requirements and identify
weaknesses in the modeling process. An evaluation agent checks each
condition in a fixed checklist. It uses the native CAD model, code, and
execution records to locate errors and examine how they are related.

To assess requirement satisfaction, we define a fixed checklist for each task.
It breaks down design requirements into checks on structure, counts,
dimensions, spatial relationships, and connectivity. For modification tasks,
the checklist covers both the
requested changes and the properties to preserve. For the same task and
input modality, all candidates and revisions use the same checklist.
Figure~\ref{fig:architecture}(b) illustrates the evaluation using a D-profile
shaft, with the transverse hole as an example of an individual requirement check.

Our evaluation agent, built on DeepSeek-V4.1-Flash, selects relevant objects
and measurements for each candidate. Requirement checks draw on native
construction records, geometric measurements from SolidWorks part (.SLDPRT)
files, and model views. For process assessment, the agent combines code,
execution logs, and actual geometry to assess the six process dimensions
defined in Sec.~\ref{sec:metrics}. It uses this evidence to trace how errors affect
subsequent modeling steps and groups issues that share a common cause
and can be resolved by the same correction.
Judgment records include the checked objects, reasons, and supporting
evidence. A scoring script verifies their completeness and consistency
before computing scores. In the illustrated example,
the agent checks the hole diameter in the PySW program and reads the
radius from native cylindrical faces. The hole measures 4.599\,mm, below the required
5.556\,mm, while the external flat meets the 2.381\,mm offset requirement.

\subsection{Evaluation Metrics}
\label{sec:metrics}

We assess modeling outcomes, processes, and revisions using the metrics
defined in Table~\ref{tab:process-metrics}.

\paragraph{Outcome.}
We report execution success rate (Exec), intersection over union (IoU),
normalized Chamfer distance (CD), and requirement satisfaction (ReqSat).
ReqSat measures how
well checklist conditions are met, including requested changes and
properties to preserve in modification tasks. Tasks without CAD output
are penalized in these scores according to Table~\ref{tab:process-metrics}.
To measure generation cost, we also report output tokens (OutTok),
averaging reasoning and answer tokens separately over recorded requests.

\input{tables/table3}

\paragraph{Process.}
We assess six dimensions across modeling strategy, 2D geometry, and 3D construction,
distinguishing requirement coverage, plan feasibility, and implementation.
At the modeling strategy stage, intent coverage (Intent) checks whether the model's
reasoning and actions cover the modeling requirements, while modeling plan
(Plan) checks whether its proposed construction approach can achieve the target.
Neither assesses implementation precision. At the 2D stage, sketch contour
(Contour) checks whether the sketch geometry can produce the target shape
under the chosen operations and placement; equivalent profiles are accepted.
Sketch parameters (Param) checks whether sketch dimensions and units meet the
requirements, geometric constraints are appropriate, and both dimensions and
constraints apply to the correct geometric elements. At the 3D stage, spatial
placement (Placement) checks whether reference planes or axes, positions,
orientations, and sides are correct. Feature construction (Feature) checks whether 3D operation types, parameters,
and end conditions are appropriate, whether operations act on the correct
objects, and whether solids are combined as required.

Each dimension is judged on its own evidence and passes only if all
applicable checks pass. Pass rates cover actual builds with CAD geometry,
including partial builds. In Figure~\ref{fig:architecture}(b), comparing the
drawing with the program reveals misread dimensions and an omitted chamfer
requirement, tracing the root cause to Intent. These misinterpretations lead
to Param errors and missing Feature, indicating difficulty reading dimensions
and local details in the engineering drawing.

To quantify error propagation, we report the conditional propagation rate
$P_{k\to l}$ (Table~\ref{tab:process-metrics}): among outputs failing
dimension $k$, the percentage with a recorded direct error-propagation
link from $k$ to $l$. Each output counts at most once per direct link.

\paragraph{Revision.}
We track ReqSat over a budget of three builds. Requirement satisfaction gain
(ReqSatGain) is the mean change from the first to the final build or between
successive steps. When a run ends early, we carry its last score forward.
Runs with no builds score zero.

Across consecutive actual builds, the error correction rate (Corr) is the
fraction of unmet conditions that become satisfied. The new error rate
(\mbox{NewErr}) is the fraction of satisfied conditions that become unmet.
The mixed revision rate (MixedRev) is the share of build pairs with
at least one correction and one new error.

We apply the same transition rates to the pass or fail verdict for each
process dimension $k$. The denominator for $\mathrm{Corr}_k$ comprises pairs
where $k$ failed in the earlier build, and that for $\mathrm{NewErr}_k$
comprises pairs where $k$ passed. Both builds in a pair must produce CAD
outputs and have process verdicts. A build without CAD output interrupts
the sequence used for pairing.

%% file: tables/table2.tex
\begin{table*}[t]
\centering
\caption{Definitions of outcome, process, and revision metrics. Token counts are in K (1,000 tokens).}
\label{tab:process-metrics}
\small
\setlength{\tabcolsep}{4pt}
\renewcommand{\arraystretch}{1.08}
\renewcommand{\tabularxcolumn}[1]{p{#1}}
\begin{tabularx}{\linewidth}{@{}
    >{\raggedright\arraybackslash}p{.10\linewidth}
    >{\raggedright\arraybackslash}p{.28\linewidth}
    >{\raggedright\arraybackslash}p{.28\linewidth}
    >{\raggedright\arraybackslash}X@{}}
\toprule
\textbf{Level} & \textbf{Metric} & \textbf{Definition} & \textbf{Description} \\
\midrule
\multirow[t]{5}{=}{\raggedright Outcome}
    & Execution success rate (Exec)
    & $\frac{1}{N}\sum_i\mathbf{1}[\mathrm{exec}_i\land\mathrm{rebuild}_i]$
    & $N$ assigned tasks. \\
    & Intersection over union (IoU)
    & $\frac{1}{N}\sum_i\begin{cases}
        \frac{|V_i\cap V_i^*|}{|V_i\cup V_i^*|}, & V_i\ne\varnothing,\\
        0, & V_i=\varnothing.
      \end{cases}$
    & $V_i/V_i^*$ candidate/reference solids. \\[2pt]
    & Normalized Chamfer distance (CD)
    & $\frac{1}{N}\sum_i\begin{cases}
        \min(1,d_i/b_i), & V_i\ne\varnothing,\\
        1, & V_i=\varnothing.
      \end{cases}$
    & $d_i$ mean bidirectional surface distance. $b_i$ reference box diagonal. \\[2pt]
    & Requirement satisfaction (ReqSat)
    & $\frac{1}{N}\sum_i\begin{cases}
        \frac{1}{K_i}\sum_j c_{ij}, & \mathrm{CAD}_i,\\
        0, & \neg\mathrm{CAD}_i.
      \end{cases}$
    & $K_i$ checklist size. $c_{ij}\in\{0,1\}$ verdicts. \\[2pt]
    & Output tokens (OutTok)
    & $\frac{1}{1000R}\sum_r(t_r^{\rm reason},t_r^{\rm answer})$
    & $R$ requests; $t_r$ raw token counts. Reasoning/answer means in K. \\
\midrule
Process
    & Intent coverage (Intent)\par
      Modeling plan (Plan)\par
      Sketch contour (Contour)\par
      Sketch parameters (Param)\par
      Spatial placement (Placement)\par
      Feature construction (Feature)
    & $\begin{aligned}[t]
      &M_k=\tfrac{1}{|G|}\textstyle\sum_{g\in G}p_{gk}\\[3pt]
      &p_{gk}=\textstyle\prod_{j\in\mathcal{C}_{gk}}c_{gj}\\[3pt]
      &c_{gj}\in\{0,1\}\\[3pt]
      &k\in\{1,\ldots,6\}
      \end{aligned}$
    & Shared scoring across six dimensions in the listed order.\par\smallskip
      $\mathcal{C}_{gk}$ applicable checks; $c_{gj}$ verdicts.\par\smallskip
      $G$ full/partial CAD outputs, one entry per actual build. \\
    & Conditional propagation rate ($P_{k\to l}$)
    & $P_{k\to l}=|G_{k\to l}^{\mathrm{link}}|/|G_k^{\mathrm{fail}}|$
    & $G_k^{\mathrm{fail}}$: outputs failing $k$.\par
      $G_{k\to l}^{\mathrm{link}}$: outputs with a direct $k\to l$ link. \\
\midrule
\multirow[t]{4}{=}{\raggedright Revision}
    & Requirement satisfaction gain (ReqSatGain)
    & $\Delta_{a\to b}=\frac{1}{N}\sum_i(s_{ib}-s_{ia})$
    & $s_{it}$ satisfaction at build $t$. Carry last score, no build $0$. \\
    & Error correction rate (Corr)
    & $(n_{U\to S})/(n_{U\to S}+n_{U\to U})$
    & \multirow{2}{=}{\raggedright $S/U$ met/unmet conditions.
        $n_{x\to y}$ adjacent-build transitions.} \\
    & New error rate (NewErr)
    & $(n_{S\to U})/(n_{S\to U}+n_{S\to S})$
    & \\
    & Mixed revision rate (MixedRev)
    & $|B_{\rm fix}\cap B_{\rm new}|/|B|$
    & $B$ adjacent-build pairs; $B_{\rm fix}/B_{\rm new}$ pairs fixing/breaking any condition. \\
\bottomrule
\end{tabularx}
\end{table*}

%% file: tables/table3.tex
\begin{table*}[!t]
\centering
\caption{Single-response generation and modification performance. Token counts are in K (1,000 tokens).}
\label{tab:main-results}
\footnotesize
\setlength{\tabcolsep}{1.35pt}
\renewcommand{\arraystretch}{1.08}
\newcommand{\vendormark}[1]{\makebox[1.05em][c]{\raisebox{-.08em}{\includegraphics[width=.92em,height=.92em,keepaspectratio]{figures/vendors/#1.png}}}\hspace{.28em}}
\newcommand{\processheadings}{Intent$\uparrow$ & Plan$\uparrow$ & Contour$\uparrow$ & Param$\uparrow$ & Placement$\uparrow$ & Feature$\uparrow$}
\begin{tabular*}{\linewidth}{@{\extracolsep{\fill}}llccccc@{\hspace{.5em}}cccccc@{}}
\toprule
 & & \multicolumn{5}{c}{Outcome} & \multicolumn{6}{c}{Process}\\
\cmidrule(lr){3-7}\cmidrule(l){8-13}
Input & Model & Exec$\uparrow$ & IoU$\uparrow$ & CD$\downarrow$ &
ReqSat$\uparrow$ & OutTok$\downarrow$ & \processheadings\\
\midrule
\multicolumn{13}{@{}l}{\itshape Generation}\\[1pt]
Text & \vendormark{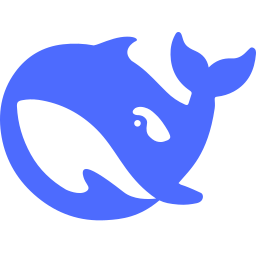}DeepSeek-V4.1-Flash & 72.67 & 56.81 & 0.2788 & 71.15 & 38.67/1.98 & 88.07 & 97.25 & 85.32 & 82.57 & \textbf{95.41} & 77.98\\
 & \vendormark{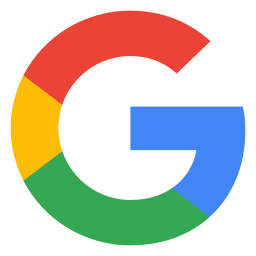}Gemini 3.8 Flash & 76.00 & 62.35 & 0.2442 & 75.13 & 38.31/1.87 & 85.96 & 95.61 & 91.23 & 88.60 & 93.86 & 78.95\\
 & \vendormark{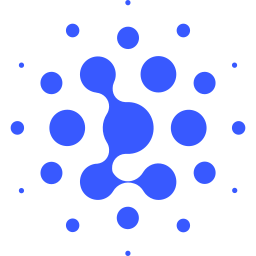}GLM-5.3-Flash & 63.33 & 50.33 & 0.3707 & 62.09 & 31.36/1.71 & 92.63 & 94.74 & 92.63 & 85.26 & 92.63 & 88.42\\
 & \vendormark{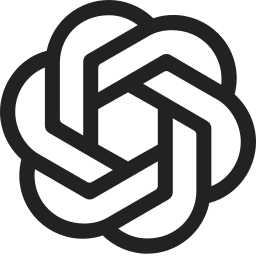}GPT-5.6-Sol & 66.00 & 54.20 & 0.3435 & 64.98 & \textbf{5.90}/\textbf{1.60} & \textbf{96.97} & 98.99 & \textbf{94.95} & \textbf{90.91} & 94.95 & \textbf{91.92}\\
 & \vendormark{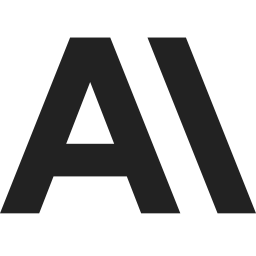}Claude Opus 5 & \textbf{86.67} & \textbf{68.04} & \textbf{0.1392} & \textbf{85.87} & 17.63/3.15 & 91.54 & \textbf{99.23} & 88.46 & 90.00 & 95.38 & 84.62\\
 & \vendormark{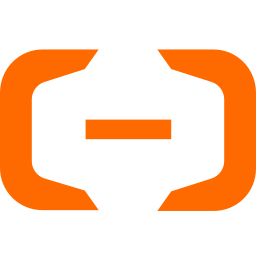}Qwen3.8-Max & 70.00 & 54.61 & 0.3055 & 68.39 & 44.49/1.81 & 92.38 & 97.14 & 88.57 & 85.71 & 92.38 & 82.86\\
 & \vendormark{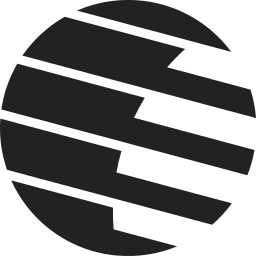}Kimi K3 & 63.33 & 48.50 & 0.3714 & 61.30 & 13.01/1.63 & 89.47 & 92.63 & 84.21 & 80.00 & 91.58 & 84.21\\
\cmidrule(lr){1-13}
Drawing & \vendormark{deepseek}DeepSeek-V4.1-Flash & 77.33 & 32.07 & 0.2506 & 33.75 & 50.37/\textbf{1.25} & 8.62 & 31.03 & 18.97 & 5.17 & 50.86 & 23.28\\
 & \vendormark{google}Gemini 3.8 Flash & 64.67 & 45.69 & 0.3602 & 49.22 & 48.97/1.84 & 34.02 & \textbf{63.92} & \textbf{44.33} & 24.74 & 76.29 & 44.33\\
 & \vendormark{zai}GLM-5.3-Flash & 60.67 & 32.63 & 0.4219 & 39.85 & 56.68/1.47 & 16.48 & 49.45 & 25.27 & 16.48 & 61.54 & 34.07\\
 & \vendormark{openai}GPT-5.6-Sol & 71.33 & 49.73 & 0.2936 & 56.47 & \textbf{7.76}/1.38 & \textbf{36.45} & 58.88 & 41.12 & 27.10 & 77.57 & \textbf{45.79}\\
 & \vendormark{anthropic}Claude Opus 5 & \textbf{85.33} & \textbf{57.87} & \textbf{0.1554} & \textbf{62.22} & 28.62/2.68 & 21.88 & 49.22 & 39.84 & 20.31 & 76.56 & 39.84\\
 & \vendormark{alibabacloud}Qwen3.8-Max & 63.33 & 40.00 & 0.3980 & 46.05 & 12.57/1.26 & 26.32 & 57.89 & 40.00 & \textbf{28.42} & \textbf{77.89} & 43.16\\
 & \vendormark{moonshot}Kimi K3 & 62.67 & 34.60 & 0.3900 & 38.95 & 27.24/1.36 & 14.89 & 44.68 & 25.53 & 17.02 & 60.64 & 35.11\\
\midrule
\multicolumn{13}{@{}l}{\itshape Modification}\\[1pt]
Code + & \vendormark{deepseek}DeepSeek-V4.1-Flash & 65.00 & 60.29 & 0.3520 & 61.35 & 40.53/2.74 & 96.15 & \textbf{100.00} & 88.46 & 96.15 & 96.15 & 84.62\\
views + & \vendormark{google}Gemini 3.8 Flash & 82.50 & 78.24 & 0.1767 & 79.45 & 36.88/2.87 & 96.97 & \textbf{100.00} & \textbf{100.00} & 90.91 & \textbf{96.97} & 90.91\\
edit text & \vendormark{zai}GLM-5.3-Flash & 67.50 & 65.31 & 0.3254 & 64.17 & 45.92/\textbf{2.23} & \textbf{100.00} & \textbf{100.00} & 92.59 & 81.48 & 96.30 & 92.59\\
 & \vendormark{openai}GPT-5.6-Sol & 72.50 & 66.15 & 0.2772 & 65.20 & \textbf{6.94}/2.53 & 96.55 & \textbf{100.00} & 86.21 & 89.66 & 89.66 & 89.66\\
 & \vendormark{anthropic}Claude Opus 5 & \textbf{92.50} & \textbf{87.83} & \textbf{0.0769} & \textbf{88.34} & 24.01/4.37 & 91.89 & 97.30 & 89.19 & 91.89 & 94.59 & 89.19\\
 & \vendormark{alibabacloud}Qwen3.8-Max & 57.50 & 53.40 & 0.4267 & 55.19 & 19.62/2.77 & 95.65 & \textbf{100.00} & 95.65 & 86.96 & 91.30 & 91.30\\
 & \vendormark{moonshot}Kimi K3 & 77.50 & 74.98 & 0.2254 & 74.06 & 22.28/2.60 & \textbf{100.00} & \textbf{100.00} & 87.10 & \textbf{96.77} & 93.55 & \textbf{93.55}\\
\bottomrule
\end{tabular*}
\par\smallskip
\begin{minipage}{\linewidth}\footnotesize
Rates and IoU are percentages; CD is normalized.
Outcomes cover all tasks; process metrics cover CAD outputs.
OutTok reports reasoning/answer tokens per request.
\end{minipage}
\end{table*}

%% file: sec/4_experiments.tex
\raggedbottom
\begin{figure*}[!t]
\centering
\begin{minipage}{\textwidth}
\input{tables/table_process_attribution}
\end{minipage}
\par\vspace{6pt}
\includegraphics[width=\textwidth]{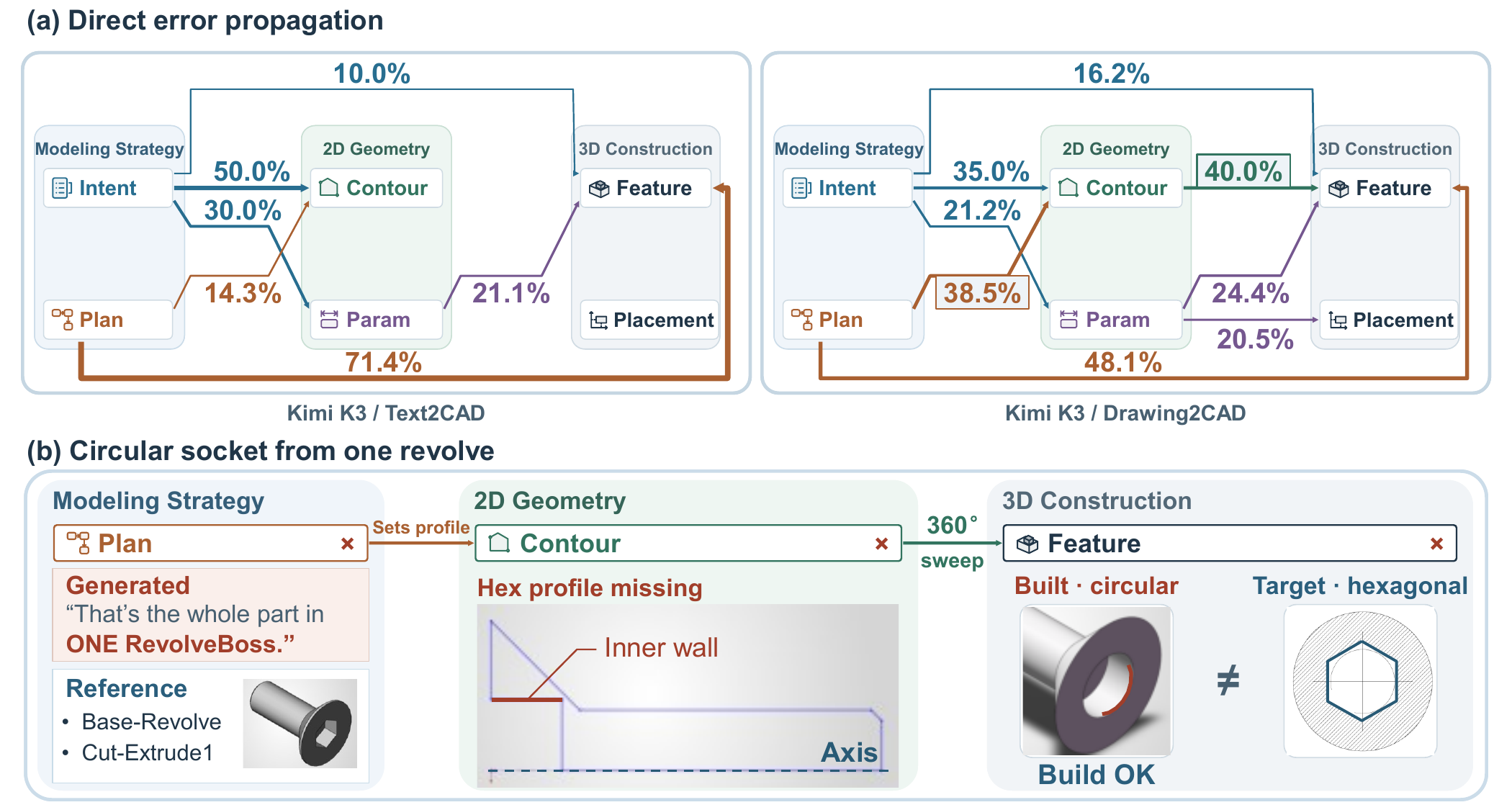}
\caption{Error propagation in Kimi K3.
(a) Direct links on Text2CAD and Drawing2CAD with conditional rates of at least 10\%.
(b) Evidence supporting a Plan$\to$Contour$\to$Feature chain:
a single-revolve plan produces a circular rather than hexagonal socket.
The drawing, model reasoning and code, native sketch, and solid geometry support the attribution.}
\label{fig:generation-propagation}
\end{figure*}

\section{Experiments}
\label{sec:experiments}

We evaluate CAD modeling across outcomes, processes, and revisions.
Beyond geometric accuracy and requirement satisfaction, we use process
metrics and error attribution records to analyze common modeling errors
and their sources across models and tasks. We then examine whether agent
revisions correct these errors or introduce new ones. Finally, we assess
the reliability of the evaluation agent.

\subsection{Experimental Setup}
\label{sec:experimental-setup}
We evaluate seven LLMs: DeepSeek-V4.1-Flash,
Gemini 3.8 Flash,
GLM-5.3-Flash, GPT-5.6-Sol,
Claude Opus 5, Qwen3.8-Max,
and Kimi K3.
Each receives the PySW API documentation and returns a complete program
in one response. Generation uses text and engineering drawings of the
same 150 common mechanical parts; modification uses
40 tasks on existing parts. Programs execute in SolidWorks.

Agent generation pairs DeepSeek-V4.1-Flash and GPT-5.6-Sol with DeepSeek
Harness (v0.1.6-alpha.1) and
Codex CLI (v0.154.0). All four configurations
use the same 40 Text2CAD tasks, documentation, and PySW inspection tools.
Each run permits at most three builds, including failed attempts. Agents
revise their own generated outputs; modification tasks instead require
editing a supplied part according to a user instruction.

\input{tables/table4}

\subsection{Main Results}
\label{sec:overall-performance}

Table~\ref{tab:main-results} shows that Claude Opus 5 achieves the highest
execution success, geometric accuracy, and requirement satisfaction across
text generation, drawing generation, and modification. All models achieve
lower requirement satisfaction on drawings than on text, highlighting the challenge of recovering
design requirements from engineering drawings. Successful execution can also
leave requirements unmet. For example, DeepSeek-V4.1-Flash achieves
77.33\% execution success but only 33.75\% requirement satisfaction on Drawing2CAD.

Process metrics reveal task-specific weaknesses. Intent coverage and
sketch parameters are the two lowest-scoring dimensions for every
model on Drawing2CAD. In modification, models score highly on intent
coverage and modeling plan, possibly benefiting from structural
similarities within part families that allow existing programs to provide
reusable modeling paths. However, weaknesses in sketch contour,
sketch parameters, and feature construction show that
reusing these paths does not ensure accurate local modifications.
These patterns guide the
following analysis of common errors, their sources, and their effects on
subsequent modeling steps.

\subsection{Modeling Process Analysis}
\label{sec:process-analysis}

Section~\ref{sec:overall-performance} shows lower requirement satisfaction
on Drawing2CAD, with intent coverage and sketch parameters as particular
weaknesses. Table~\ref{tab:generation-attribution} connects these failures:
all models record Intent$\to$Param and Intent$\to$Contour links on Drawing2CAD,
indicating that some dimensional and contour errors stem from misinterpreting
design requirements. Geometric deviations can therefore originate in
requirement interpretation.

Errors also propagate from modeling plans and sketch contours to feature
construction. In the aggregate results of Table~\ref{tab:generation-attribution},
more than half of the outputs failing modeling plan record a
Plan$\to$Feature link in both tasks; Contour$\to$Feature is also prominent
on Drawing2CAD. For Kimi, Figure~\ref{fig:generation-propagation}(a) shows
Intent$\to$Contour, Plan$\to$Contour, and Contour$\to$Feature links,
identifying both upstream sources of contour errors and their downstream
effects. Analyzing feature-level deviations thus requires examining earlier
requirement interpretation, planning, and sketch construction.

Figure~\ref{fig:generation-propagation}(a) summarizes recurring direct links
across outputs, while Figure~\ref{fig:generation-propagation}(b) shows how
errors develop along a chain within one output. Although the drawing requires a hexagonal socket, Kimi's
reasoning and code adopt a single full revolution to construct the entire
part. This plan represents the hole through an axial half-profile, whose
inner wall revolves into a cylindrical surface, producing a circular hole
instead of a hexagonal socket. The program executes successfully, but the
chosen plan cannot represent the required non-axisymmetric shape; adjusting
the hole radius or depth alone cannot resolve the error. This example grounds
the recorded Plan$\to$Contour$\to$Feature chain in a specific modeling
decision, showing how process evaluation identifies both the final defect
and the earlier step that requires revision.

\subsection{Agent Performance}
\label{sec:agent-performance}

\input{tables/table5}

The preceding analysis identifies weaknesses in modeling and how errors
propagate. This section evaluates whether different model--harness pairings can
use tool feedback to improve modeling outcomes while preserving correct
results.

Table~\ref{tab:agent-detailed}(a) shows that all four configurations improve
requirement satisfaction, but the better-performing harness depends on the
model: DeepSeek-V4.1-Flash performs better with DeepSeek Harness, whereas
GPT-5.6-Sol performs better with Codex CLI. For GPT-5.6-Sol, Codex CLI
outperforms DeepSeek Harness in both first-build requirement satisfaction
and revision gain. For DeepSeek-V4.1-Flash, DeepSeek Harness starts slightly
lower but achieves a better final result through larger revision gains.
Model--harness compatibility thus affects both initial generation and
subsequent improvement using feedback. Table~\ref{tab:agent-detailed}(b)
locates these differences within the modeling process. For GPT-5.6-Sol, Codex CLI
has higher failure-to-pass rates and lower pass-to-failure rates across
all six dimensions. For DeepSeek-V4.1-Flash, DeepSeek Harness more often improves
intent coverage, modeling plans, sketch parameters, spatial placement,
and feature construction, while Codex CLI better preserves correct spatial
placement. Model--harness compatibility thus affects both final performance
and the balance between correction and preservation.

Some aspects of modeling are harder to correct than others, and revisions
can also break previously correct results. In Table~\ref{tab:agent-detailed}(b),
Feature has the lowest failure-to-pass rate in all four configurations,
identifying feature construction as a shared revision difficulty. Sketch parameters
can also regress during revision: among consecutive-build pairs where Param
initially passes, at least 19.4\% subsequently fail in every configuration.
Correct dimensions, constraints, and object bindings therefore remain
vulnerable to revision. MixedRev captures such tradeoffs within a single
revision: despite its low condition-level new error rate, GPT-5.6-Sol with Codex CLI
both fixes and introduces errors in 27.8\% of revisions
(Table~\ref{tab:agent-detailed}(a)). Strong aggregate gains can thus coexist
with local regressions. Together, Corr, NewErr, MixedRev, and dimension-level
transitions reveal how often errors are corrected, whether corrections
coincide with new errors, and which aspects of modeling improve or regress.

\subsection{Evaluator Validation}
\label{sec:evaluator-reliability}

We compare PySWBench with adapted implementations of
BlenderLLM~\citep{du2024blenderllm}, CADTests~\citep{mallis2026text},
and MUSE~\citep{dong2026muse} on the same candidates and acceptance
checklists. The evaluation includes 84 candidate parts sampled across
difficulty levels, covering 1,642 conditions, and 23 candidates with
challenging-to-detect errors, covering 516 conditions. Each experiment is
repeated three times. Evaluator correctness, misses, and false alarms are
assessed against human-annotated gold-standard labels.

For $N$ conditions, let $D$ be those judged to satisfy or violate
requirements,
$C$ those judged correctly, $V$ the reference violations, and $F$ those
judged to violate requirements. We define coverage as $\mathrm{Cov}=|D|/N$, accuracy as
$\mathrm{Acc}=|C|/N$, false negative rate as
$\mathrm{FNR}=|V\setminus F|/|V|$, and false positive rate as
$\mathrm{FPR}=|F\setminus V|/(N-|V|)$.
Cov measures how often the evaluator reaches a decision; Acc
measures how often it is correct. FNR and FPR quantify
undetected violations and incorrectly rejected satisfied conditions,
respectively. Conditions without a definite judgment reduce coverage and
accuracy, and count as misses when the reference indicates a violation. On $D$, Cohen's
$\kappa$ measures agreement beyond chance:
\begin{equation}
\kappa=\frac{p_o-p_e}{1-p_e},\qquad
p_o=\frac{|C|}{|D|},\qquad
p_e=\sum_{y\in\{0,1\}}r_yq_y,
\end{equation}
where $r_y$ and $q_y$ are the reference and predicted proportions of label
$y$ within $D$, where 0 denotes satisfaction and 1 denotes violation.

PySWBench achieves full decision coverage, the highest $\kappa$ and accuracy,
and the lowest miss and false-alarm rates on both sets. The stress set
highlights these advantages: CADTests has a relatively similar miss rate
but substantially more false alarms, partly because fixed test rules fail
to locate the intended structures and mistake these mismatches for
requirement violations. MUSE has a lower false-alarm rate than CADTests,
but also lower coverage and a higher miss rate. PySWBench therefore offers
a better balance between detecting violations and accepting correct
results. However, its stress-set miss rate remains 31.21\%, leaving room
to improve the verification of complex conditions.

%% file: tables/table_process_attribution.tex
\begingroup
\centering
\captionof{table}{Conditional error propagation in end-to-end generation (\%).}
\label{tab:generation-attribution}
\footnotesize
\setlength{\tabcolsep}{1.5pt}
\renewcommand{\arraystretch}{1.08}
\begin{tabular*}{\textwidth}{@{\extracolsep{\fill}}l*{10}{c}@{}}
\toprule
 & \multicolumn{2}{c}{$P_{\mathrm{Contour}\to\mathrm{Feature}}$} & \multicolumn{2}{c}{$P_{\mathrm{Plan}\to\mathrm{Feature}}$} & \multicolumn{2}{c}{$P_{\mathrm{Intent}\to\mathrm{Contour}}$} & \multicolumn{2}{c}{$P_{\mathrm{Param}\to\mathrm{Feature}}$} & \multicolumn{2}{c}{$P_{\mathrm{Intent}\to\mathrm{Param}}$} \\
\cmidrule(lr){2-3}\cmidrule(lr){4-5}\cmidrule(lr){6-7}\cmidrule(lr){8-9}\cmidrule(l){10-11}
Model & Text & Drawing & Text & Drawing & Text & Drawing & Text & Drawing & Text & Drawing \\
\midrule
DeepSeek-V4.1-Flash & 12.5 & 51.1 & 100.0 & 57.5 & 38.5 & 26.4 & 5.3 & 27.3 & 23.1 & 18.9 \\
Gemini 3.8 Flash & 40.0 & 37.0 & 80.0 & 57.1 & 6.2 & 21.9 & 0.0 & 23.3 & 6.2 & 21.9 \\
GLM-5.3-Flash & 0.0 & 35.3 & 60.0 & 43.5 & 42.9 & 27.6 & 14.3 & 21.1 & 0.0 & 22.4 \\
GPT-5.6-Sol & 40.0 & 38.1 & 100.0 & 43.2 & 33.3 & 22.1 & 0.0 & 24.4 & 33.3 & 16.2 \\
Claude Opus 5 & 20.0 & 40.3 & 100.0 & 53.8 & 36.4 & 23.0 & 0.0 & 15.7 & 0.0 & 27.0 \\
Qwen3.8-Max & 16.7 & 33.3 & 66.7 & 52.5 & 37.5 & 27.1 & 0.0 & 22.1 & 12.5 & 21.4 \\
Kimi K3 & 6.7 & 40.0 & 71.4 & 48.1 & 50.0 & 35.0 & 21.1 & 24.4 & 30.0 & 21.2 \\
\midrule
\textbf{Overall} & \textbf{17.5} & \textbf{40.2} & \textbf{76.0} & \textbf{51.4} & \textbf{32.4} & \textbf{26.2} & \textbf{6.9} & \textbf{22.6} & \textbf{13.2} & \textbf{21.5} \\
\bottomrule
\end{tabular*}
\par\smallskip
\begin{minipage}{\linewidth}\footnotesize
Rates are conditioned on failure in the source dimension; each output counts once per direct link.
\end{minipage}
\endgroup

%% file: tables/table4.tex
\begin{table*}[t]
\centering
\caption{Agent outcomes and process changes on Text2CAD tasks. Token counts are in K (1,000 tokens).}
\label{tab:agent-detailed}
\footnotesize
\setlength{\tabcolsep}{2pt}
\renewcommand{\arraystretch}{1.12}
\noindent\textbf{(a) Outcomes and revision behavior}\par\smallskip
\begin{tabular*}{\linewidth}{@{\extracolsep{\fill}}ll*{10}{c}@{}}
\toprule
 & & \multicolumn{5}{c}{Final build and task usage} & \multicolumn{5}{c}{Revision metrics}\\
\cmidrule(lr){3-7}\cmidrule(l){8-12}
Model & Harness & Exec$\uparrow$ & IoU$\uparrow$ & CD$\downarrow$ &
ReqSat$\uparrow$ & Tok$\downarrow$ & ReqSat$_1\uparrow$ & ReqSatGain$\uparrow$ & Corr$\uparrow$ & NewErr$\downarrow$ & MixedRev\\
\midrule
DeepSeek-V4.1-Flash & DeepSeek Harness & 80.00 & 60.37 & .0371 & 78.48 & 1178.5 & 31.49 & \textbf{46.99} & \textbf{60.18} & \textbf{3.94} & 8.9\\
 & Codex CLI & 75.00 & 59.22 & .0156 & 73.34 & 2074 & 35.83 & 37.51 & 56.60 & 6.28 & 14.8\\
GPT-5.6-Sol & DeepSeek Harness & 62.50 & 59.64 & .0146 & 55.25 & \textbf{255} & 22.01 & 33.24 & 35.68 & 11.69 & 12.5\\
 & Codex CLI & \textbf{85.00} & \textbf{62.69} & \textbf{.0122} & \textbf{79.60} & 663 & \textbf{37.52} & 42.08 & 45.87 & 3.98 & 27.8\\
\bottomrule
\end{tabular*}
\par\medskip
\noindent\textbf{(b) Process changes: $\mathrm{Corr}_k\uparrow$ / $\mathrm{NewErr}_k\downarrow$ (\%)}\par\smallskip
\begin{tabular*}{\linewidth}{@{\extracolsep{\fill}}ll*{6}{c}@{}}
\toprule
Model & Harness & Intent & Plan & Contour & Param & Placement & Feature\\
\midrule
DeepSeek-V4.1-Flash & DeepSeek Harness & 65.0 / 11.1 & 75.0 / 0.0 & 61.9 / 11.4 & 50.0 / 19.4 & 73.1 / 30.0 & 35.1 / 36.8\\
 & Codex CLI & 53.8 / 8.3 & 71.4 / 7.4 & 63.6 / 7.7 & 40.0 / 19.5 & 60.0 / 8.3 & 30.2 / 44.4\\
GPT-5.6-Sol & DeepSeek Harness & 25.0 / 35.3 & 31.0 / 12.5 & 29.0 / 31.8 & 17.9 / 28.0 & 20.5 / 66.7 & 10.6 / 83.3\\
 & Codex CLI & 50.0 / 23.3 & 69.2 / 9.8 & 45.5 / 18.8 & 41.4 / 24.0 & 30.8 / 20.0 & 22.7 / 40.0\\
\bottomrule
\end{tabular*}
\par\smallskip
\begin{minipage}{\linewidth}\footnotesize
(a) Rates and IoU are percentages; ReqSatGain is in percentage points.
ReqSat$_1$ denotes first-build satisfaction; Corr/NewErr are condition-level rates.
Tok counts input and output tokens per task, including cached input.
All outcome metrics cover 40 tasks.
(b) Cells report dimension-level failure-to-pass / pass-to-failure rates for consecutive builds with CAD outputs at both endpoints.
\end{minipage}
\end{table*}

%% file: tables/table5.tex
\begin{table*}[t]
\centering
\caption{Evaluator reliability on individual conditions. Entries are means $\pm$ sample standard deviations over three trials; all metrics except $\kappa$ are percentages.}
\label{tab:evaluator-reliability}
\footnotesize
\setlength{\tabcolsep}{4pt}
\renewcommand{\arraystretch}{1.12}
\begin{tabular*}{\textwidth}{@{\extracolsep{\fill}}lccccc@{}}
\toprule
Evaluator & Cov$\uparrow$ & $\kappa\uparrow$ & Acc$\uparrow$ & FNR$\downarrow$ & FPR$\downarrow$\\
\midrule
\multicolumn{6}{@{}l}{\textit{(a) Difficulty-stratified sample: 1,642 conditions, 39 violations}}\\[2pt]
BlenderLLM & $\mathbf{100.00}\,\pm\,0.00$ & $0.736\,\pm\,0.038$ & $98.74\,\pm\,0.31$ & $24.79\,\pm\,5.34$ & $0.69\,\pm\,0.45$\\
CADTests & $95.82\,\pm\,3.93$ & $0.145\,\pm\,0.027$ & $80.67\,\pm\,1.96$ & $31.62\,\pm\,18.19$ & $15.03\,\pm\,4.17$\\
MUSE & $56.07\,\pm\,0.79$ & $0.311\,\pm\,0.171$ & $54.47\,\pm\,0.94$ & $82.91\,\pm\,10.36$ & $1.23\,\pm\,0.24$\\
PySWBench & $\mathbf{100.00}\,\pm\,0.00$ & $\mathbf{0.850}\,\pm\,0.032$ & $\mathbf{99.35}\,\pm\,0.14$ & $\mathbf{20.51}\,\pm\,2.56$ & $\mathbf{0.17}\,\pm\,0.10$\\
\midrule
\multicolumn{6}{@{}l}{\textit{(b) High-detection-difficulty stress set: 516 conditions, 94 violations}}\\[2pt]
BlenderLLM & $99.55\,\pm\,0.49$ & $0.506\,\pm\,0.133$ & $86.56\,\pm\,2.64$ & $49.65\,\pm\,13.13$ & $5.13\,\pm\,0.76$\\
CADTests & $88.05\,\pm\,0.78$ & $0.386\,\pm\,0.124$ & $66.60\,\pm\,4.96$ & $36.52\,\pm\,13.72$ & $23.46\,\pm\,4.04$\\
MUSE & $48.77\,\pm\,0.30$ & $0.278\,\pm\,0.016$ & $39.99\,\pm\,0.49$ & $85.11\,\pm\,2.81$ & $3.87\,\pm\,0.60$\\
PySWBench & $\mathbf{100.00}\,\pm\,0.00$ & $\mathbf{0.762}\,\pm\,0.005$ & $\mathbf{93.67}\,\pm\,0.11$ & $\mathbf{31.21}\,\pm\,2.21$ & $\mathbf{0.79}\,\pm\,0.55$\\
\bottomrule
\end{tabular*}

\end{table*}

%% file: sec/5_conclusion.tex
\section{Conclusion}
\label{sec:conclusion}

PySWBench combines native CAD data, tools, and agentic evaluation to verify
design requirements and analyze modeling errors. Experiments reveal shared weaknesses
in drawing interpretation and dimension setting, with upstream errors affecting
later construction. Feedback-based revisions improve results but can introduce
new errors, highlighting the need to correct failures while preserving satisfied
requirements. Comparisons with human annotations support the evaluator's
reliability. By connecting requirement checks with evidence from the modeling
process, PySWBench identifies both final defects and earlier decisions that
require revision, supporting more targeted improvements to CAD generation.
However, complex violations remain difficult to detect, leaving room to improve
the evaluator's verification capabilities.

%% file: sec/appendices.tex
\onecolumn
\setlength{\multicolsep}{5pt}
\setlength{\columnsep}{36pt}
\appendix
\section{Benchmark Design}
\input{sec/supplementary/1_modeling_interface}
\input{sec/supplementary/3_corpus}
\begin{multicols}{2}
\input{sec/supplementary/4_tasks_checklists}
\end{multicols}
\clearpage
\section{Evaluator Design and Validation}
\input{sec/supplementary/11_evaluator_evidence}
\input{sec/supplementary/12_baseline_adaptation}

\input{sec/supplementary/13_process_validation}
\input{sec/supplementary/14_annotation_records}
\input{sec/supplementary/11_freecad}
\clearpage
\section{Generation Performance by Complexity}
\input{sec/supplementary/15_complexity_trends}
\input{sec/supplementary/16_complexity_factors}
\input{sec/supplementary/17_reproducibility}

%% file: sec/supplementary/1_modeling_interface.tex
\subsection{PySW Modeling and Inspection}
\label{app:api}
\label{app:inspection}

\begin{multicols}{2}
PySW exposes SolidWorks operations through a Python COM interface.
Table~\ref{tab:pysw-api-capabilities} summarizes its modeling, drawing,
and inspection tools. PySW reads native construction
records, measures geometry, captures model and edit-mode views, and
reports rebuild errors. It also exports reference-part records as JSON
and validates the reference packages.

\end{multicols}
\begin{table}[H]
\centering
\caption{PySW operations for part modeling, engineering drawings, and native CAD inspection.}
\label{tab:pysw-api-capabilities}
\label{tab:inspection-tools}
\setlength{\tabcolsep}{7pt}
\renewcommand{\arraystretch}{1.15}
\begin{tabularx}{\textwidth}{@{}>{\raggedright\arraybackslash}p{0.16\textwidth}
>{\raggedright\arraybackslash}p{0.18\textwidth}
>{\raggedright\arraybackslash}X@{}}
\toprule
\textbf{Group} & \textbf{Category} & \textbf{Capabilities} \\
\midrule
\textbf{Sketching} & Geometry and editing &
2D and 3D sketch entities, including lines, arcs, conics, and splines.
Editing tools include projection, trimming, and 2D offsets and patterns. \\
& Constraints and dimensions &
Geometric relations such as coincidence, tangency, symmetry, and fixed geometry.
Dimensions include linear, projected, and doubled distances, radii, diameters,
arc lengths, and angles. \\
\addlinespace[5pt]
\textbf{Part features} & Reference geometry and curves &
Planes, axes, points, coordinate systems, helices, composite and projected curves,
curves through 3D points, and split lines \\
& Solids and surfaces &
Boss and cut features (extrusions, revolutions, sweeps, lofts, and boundary
features), surface lofts, filled surfaces, and thickening \\
& Shape modification &
Fillets, chamfers, shells, drafts, domes, ribs, and wraps \\
& Patterns and mirrors &
Linear and circular patterns and mirroring of features or bodies \\
& Bodies and derived parts &
Boolean combinations and intersections, Move/Copy Body, Delete/Keep Body,
and Insert Part (derived parts) \\
\addlinespace[5pt]
\textbf{Drawings} & Detailing and export &
Standard and named drawing views, dimensions, notes, and PDF export \\
\addlinespace[5pt]
\textbf{Modeling support} & Documents and state &
SolidWorks sessions, part creation, opening, closing, and saving, and
feature suppression and unsuppression by configuration \\
& Topology selection &
Face, edge, and body references by geometric property, feature, or sketch profile \\
& Equations &
Global variables, equations, driving-dimension bindings, equation evaluation,
and dimension queries \\
\addlinespace[5pt]
\textbf{Model inspection} & Model diagnostics &
Rebuild errors and body faults \\
& Construction records &
Feature trees, definitions, dependencies, and parameters, plus sketch entities,
geometric relations, and driving and driven dimensions \\
& Geometry queries &
Body bounds, volume, surface area, analytic face and edge geometry and its
parameters, distances, diameters, angles, counts, and derived quantities \\
& Visual inspection &
Model views, sections, entity highlights, and sketch and feature edit-state screenshots \\
\bottomrule
\end{tabularx}
\end{table}

%% file: sec/supplementary/3_corpus.tex
\clearpage
\subsection{Construction Complexity}
\label{app:construction}
\label{app:dataset-complexity}

\begin{multicols}{2}
We compute construction complexity from the native SolidWorks feature
histories of the 150 reference parts shared by Text2CAD and Drawing2CAD.
The score combines three counts.

\smallskip
\noindent\textbf{Solid features ($F_i$).}
We count each solid modeling feature once, including extrusions, cuts,
fillets, and patterns. A pattern contributes one feature regardless of
the number of instances. The count excludes sketches, duplicate tree
references, and auxiliary entries such as helices, cosmetic threads,
design tables, weldment and sheet-metal definitions, and flat patterns.

\smallskip
\noindent\textbf{Sketch-plane changes ($P_i$).}
We compare the support planes of consecutive 2D sketches in feature-tree
order. A change in orientation or a parallel offset counts once. Opposite
normals on the same plane are equivalent. We compare unit normals and
signed offsets with tolerances of $10^{-5}$ and $1\,\mu\mathrm{m}$,
respectively. A 3D sketch interrupts the sequence.

\smallskip
\noindent\textbf{Maximum sketch size ($S_i$).}
We count the lines, arcs, and other curve segments in each 2D or 3D sketch
and take the largest count. Construction geometry and standalone points
are excluded. For parts without sketches, $S_i=0$.

\noindent To compute the score, we rank the $M=150$ reference parts in
ascending order for each count, assigning average ranks to ties. We
combine the normalized ranks on a 0--100 scale.
\begin{align}
R(x_i) &= \frac{\operatorname{rank}_{\mathrm{avg}}(x_i)-1}{M-1},
          \quad x\in\{F,P,S\}, \label{eq:normalized-rank}\\
K_i^{\mathrm{comp}} &= 100\bigl[0.2R(F_i)+0.4R(P_i)+0.4R(S_i)\bigr].
          \label{eq:construction-complexity}
\end{align}

\begin{figure}[H]
\centering
\includegraphics[width=.62\textwidth]{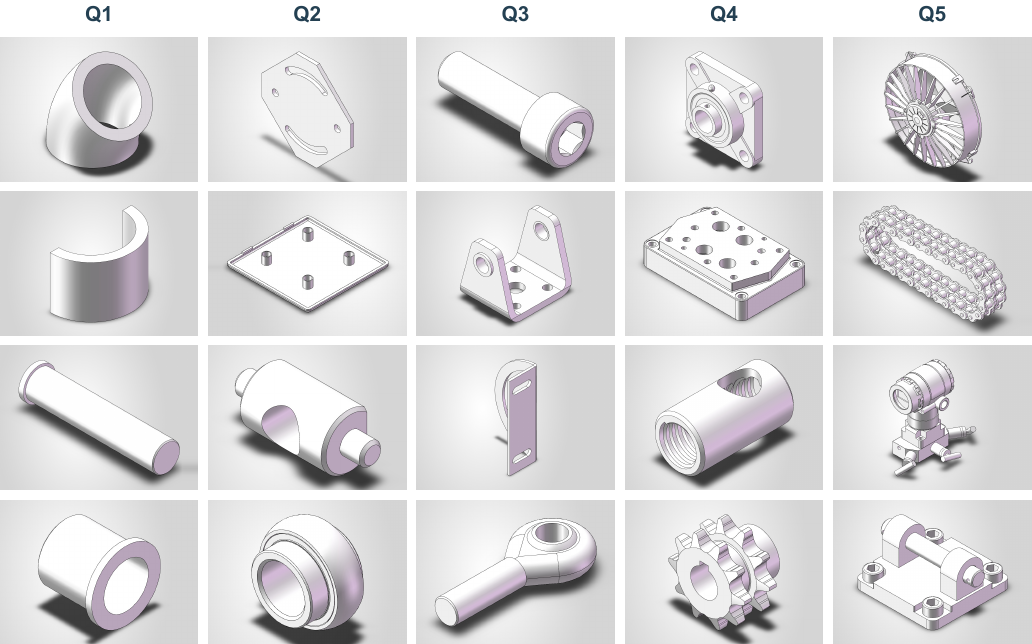}
\caption{SolidWorks views of four reference parts from each construction
complexity quintile, ordered from Q1 (lowest) to Q5 (highest).}
\label{fig:reference-parts}
\end{figure}

We group parts into quintiles and keep tied scores together.
Table~\ref{tab:reference-complexity} summarizes each quintile.
Figure~\ref{fig:reference-parts} shows four parts per quintile.

\end{multicols}
\begin{table}[H]
\centering
\caption{Construction complexity of 150 reference parts. Each row gives
the score range and median factor counts within a quintile.}
\label{tab:reference-complexity}
\small
\setlength{\tabcolsep}{4pt}
\begin{tabular*}{.75\linewidth}{@{\extracolsep{\fill}}lccrrr@{}}
\toprule
\multicolumn{3}{@{}l}{} & \multicolumn{3}{c@{}}{Median count} \\
\cmidrule(l){4-6}
Group & Parts & Complexity range & $F$ & $P$ & $S$ \\
\midrule
Q1 & 32 & 6.51--31.07  &  1 & 0 & 5.5 \\
Q2 & 29 & 31.88--42.21 &  3 & 1 & 8 \\
Q3 & 29 & 42.68--55.37 &  5 & 3 & 8 \\
Q4 & 30 & 55.70--69.13 &  8 & 3 & 12 \\
Q5 & 30 & 71.01--98.26 & 14 & 7 & 24 \\
\bottomrule
\end{tabular*}
\end{table}

%% file: sec/supplementary/4_tasks_checklists.tex
\clearpage
\subsection{Acceptance Checklists}
\label{app:tasks}
\input{sec/supplementary/checklist_view_overlays}

We group acceptance conditions by feature or part region. Each condition
has an ID and records the object, property, acceptance criterion, and
source requirement. Modification checklists cover requested edits and
properties to preserve.

\noindent Table~\ref{tab:checklist-example} lists 22 generation
checks for the twin-lug support across the regions marked in
Figure~\ref{fig:checklist-support}.

\begingroup
\setlength{\intextsep}{6pt}
\captionsetup{skip=6pt}
\begin{figure}[H]
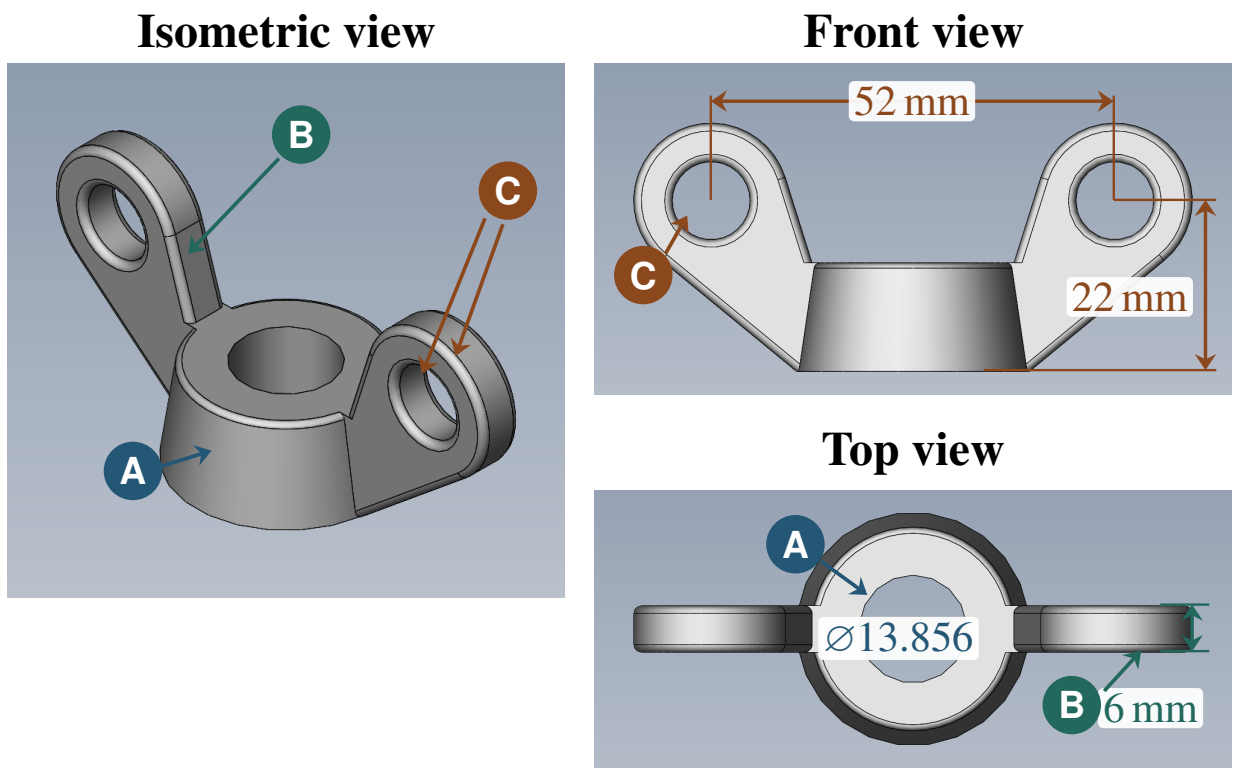

\centering
\checksupportviews
\caption{Checklist locations on the twin-lug support. Dimensions are in
millimeters. \checktag{checkblue}{A}~Body and central bore.
\checktag{checkteal}{B}~Lug plates.
\checktag{checkrust}{C}~Lug bores and blends.}
\label{fig:checklist-support}
\end{figure}

\begin{table}[H]
\centering
\caption{Acceptance checklist for the twin-lug support in Text2CAD.}
\label{tab:checklist-example}
\small
\setlength{\tabcolsep}{4pt}
\renewcommand{\arraystretch}{1.0}
\begin{tabularx}{\columnwidth}{@{}l>{\raggedright\arraybackslash}X@{}}
\toprule
ID & Acceptance condition \\
\midrule
\multicolumn{2}{@{}l}{\checktag{checkblue}{A}\quad\textit{Part and annular body}} \\
R01.01 & Twin-lug annular support. \\
R01.02 & One connected solid. \\
R01.03 & Envelope $\approx72\times32\times30$\,mm. \\
R02.01 & Frustum-shaped annular body. \\
R02.02 & Central through bore. \\
R02.03 & Base outside diameter of 30\,mm. \\
R02.04 & Extrapolated cone $\varnothing$26\,mm at 14\,mm above base. \\
R02.05 & Axial body height of 14\,mm. \\
R02.06 & Central bore diameter of 13.856\,mm. \\
\addlinespace[2pt]
\multicolumn{2}{@{}l}{\checktag{checkteal}{B}\quad\textit{Lug plates}} \\
R03.01 & Two lugs extend outward and up from opposite sides. \\
R03.02 & Lugs mirror about the midplane through the body axis. \\
R03.03 & Each lug is 6\,mm thick. \\
R03.05 & Both straight lug sides are tangent to the $R10$ head. \\
R03.06 & Straight sides join the top and bottom body edges. \\
R05.02 & Both lugs are continuously joined to the body. \\
\addlinespace[2pt]
\multicolumn{2}{@{}l}{\checktag{checkrust}{C}\quad\textit{Lug heads, bores and blends}} \\
R03.04 & Lug-bore centers are 22\,mm above the base. \\
R04.01 & Each lug ends in an arc of radius 10\,mm. \\
R04.02 & Each lug head has one $\varnothing$10\,mm through bore. \\
R04.03 & Lug-bore axes are parallel and distinct. \\
R04.04 & Lug bores have a center spacing of 52\,mm. \\
R05.01 & Central bore axis perpendicular to both lug-bore axes. \\
R05.03 & Blend outer lug edges and both rims of each lug bore. \\
\bottomrule
\end{tabularx}
\end{table}
\endgroup

\columnbreak

\begin{figure}[H]
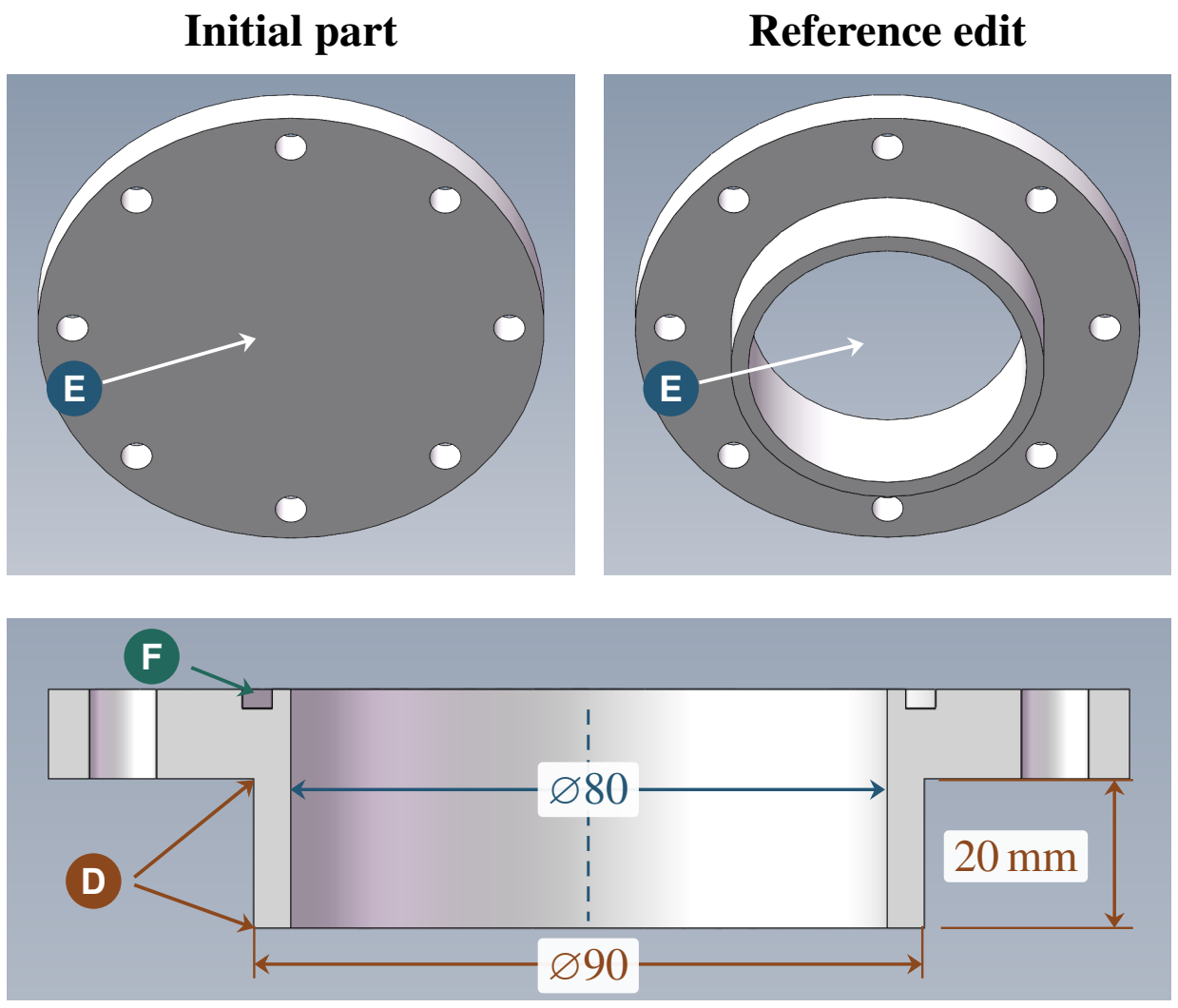

\centering
\checkflangeviews
\caption{Checklist locations on the blind flange before and after modification. The
axial section shows the reference edit. \checktag{checkrust}{D}~Added neck
and sharp edges. \checktag{checkblue}{E}~Closed face and through bore.
\checktag{checkteal}{F}~Sealing groove to preserve.}
\label{fig:checklist-flange}
\end{figure}

\noindent Figure~\ref{fig:checklist-flange} shows the blind flange modification,
which adds a coaxial neck and a bore through the neck and flange.
Table~\ref{tab:modification-checklist-example} also specifies which
geometry to preserve.

\begin{table}[H]
\centering
\caption{Acceptance checklist for the blind flange modification.}
\label{tab:modification-checklist-example}
\small
\setlength{\tabcolsep}{5pt}
\renewcommand{\arraystretch}{1.15}
\begin{tabularx}{\columnwidth}{@{}l>{\raggedright\arraybackslash}X@{}}
\toprule
Aspect & Acceptance condition \\
\midrule
\multicolumn{2}{@{}l}{\checktag{checkrust}{D}\quad\textit{Added neck}} \\
Datum      & The neck starts at the original ungrooved rear face. \\
Alignment  & The neck extends outward, coaxial with the original flange. \\
Length     & The neck extends 20\,mm from the original rear face. \\
Neck diameter & The neck has an outside diameter of 90\,mm. \\
Neck mouth & The edge at the neck mouth is sharp. \\
Junction   & The edge at the junction step is sharp. \\
\addlinespace[2pt]
\multicolumn{2}{@{}l}{\checktag{checkblue}{E}\quad\textit{Through bore}} \\
Bore diameter & The central bore has a diameter of 80\,mm. \\
Bore extent & The bore runs through both the flange and the neck. \\
Removal    & The original face that closed the center is absent. \\
\addlinespace[2pt]
\multicolumn{2}{@{}l}{\checktag{checkteal}{F}\quad\textit{Retained geometry}} \\
Preservation & The original geometry outside the requested edit is
               unchanged. \\
\bottomrule
\end{tabularx}
\end{table}

%% file: sec/supplementary/checklist_view_overlays.tex
\definecolor{checkblue}{HTML}{245675}
\definecolor{checkteal}{HTML}{22685C}
\definecolor{checkrust}{HTML}{8C481D}
\tikzset{
  checkbadge/.style={circle, inner sep=0pt, minimum size=11pt,
    text=white, font=\sffamily\bfseries\scriptsize},
  checklabel/.style={fill=white, fill opacity=.94, text opacity=1,
    rounded corners=1pt, inner sep=2pt, font=\small},
  checkline/.style={line width=.7pt, -{Stealth[length=3pt,width=4pt]}},
  checkdim/.style={line width=.65pt,
    {Stealth[length=3pt,width=4pt]}-{Stealth[length=3pt,width=4pt]}}
}

\newcommand{\checktag}[2]{\textcolor{#1}{\sffamily\textbf{#2}}}
\newsavebox{\checkviewport}
\newcommand{\checkview}[3]{%
  \sbox{\checkviewport}{\includegraphics[width=#1]{figures/checklist-#2.png}}%
  \begin{tikzpicture}[baseline=(view.south),x=\wd\checkviewport,
      y=-\ht\checkviewport]
    \node[inner sep=0pt,anchor=north west] (view) at (0,0) {\usebox{\checkviewport}};
    #3%
  \end{tikzpicture}%
}

\newcommand{\checksupportviews}{%
  \begin{minipage}[t]{.455\columnwidth}
  \centering
  \small\textbf{Isometric view}\par\vspace{2pt}
  \checkview{\linewidth}{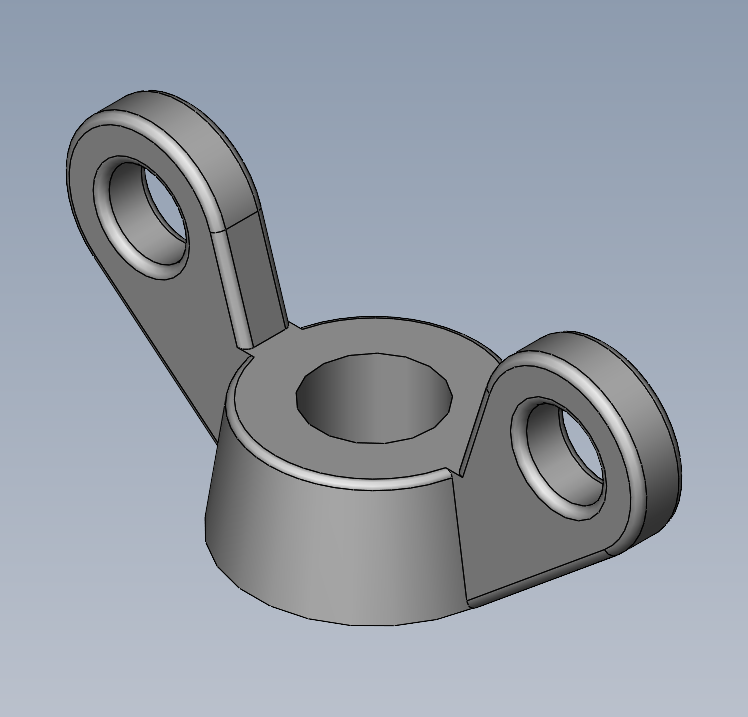}{%
    \node[checkbadge,fill=checkblue] at (.228,.765) {A};
    \draw[checkline,checkblue] (.271,.759) -- (.369,.727);
    \node[checkbadge,fill=checkteal] at (.53,.135) {B};
    \draw[checkline,checkteal] (.49,.180) -- (.327,.358);
    \node[checkbadge,fill=checkrust] at (.901,.240) {C};
    \draw[checkline,checkrust] (.892,.303) -- (.807,.55);
    \draw[checkline,checkrust] (.86,.292) -- (.747,.583);
  }%
  \end{minipage}\hfill
  \begin{minipage}[t]{.52\columnwidth}
  \centering
  \small\textbf{Front view}\par\vspace{2pt}
  \checkview{\linewidth}{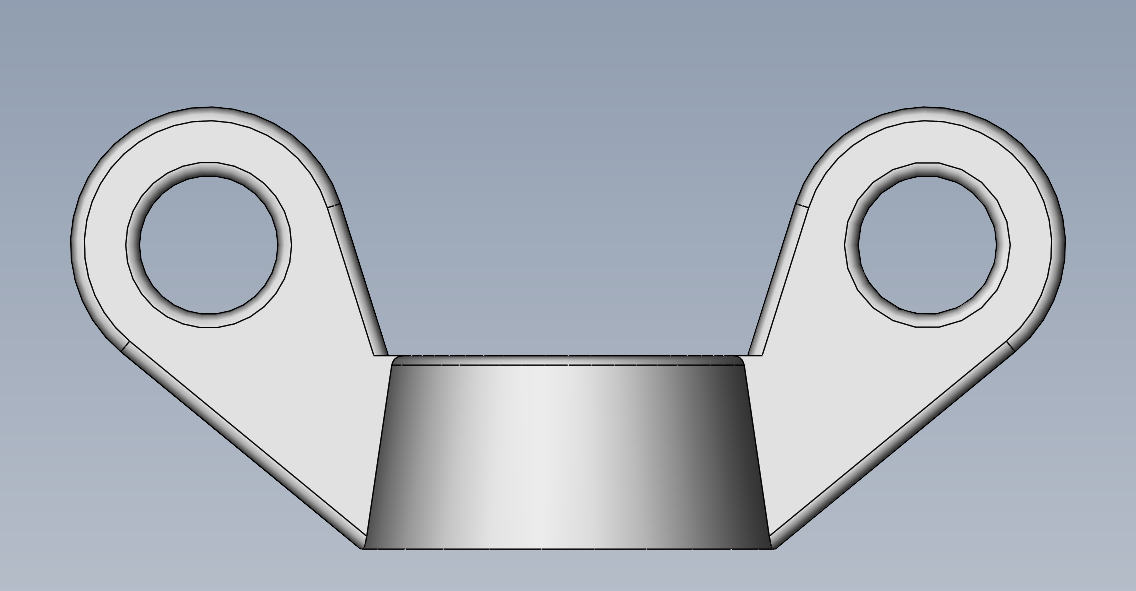}{%
    \draw[checkdim,checkrust] (.184,.117) -- (.817,.117);
    \draw[checkrust,line width=.45pt] (.184,.1) -- (.184,.415)
                                       (.817,.1) -- (.817,.415);
    \node[checklabel,font=\footnotesize,text=checkrust,inner sep=1pt]
      at (.50,.117) {52\,mm};
    \draw[checkdim,checkrust] (.961,.415) -- (.961,.929);
    \draw[checkrust,line width=.45pt] (.817,.415) -- (.978,.415)
                                      (.613,.929) -- (.978,.929);
    \node[checklabel,font=\footnotesize,text=checkrust,anchor=east,inner sep=1pt]
      at (.943,.708) {22\,mm};
    \node[checkbadge,fill=checkrust] at (.078,.642) {C};
    \draw[checkline,checkrust] (.113,.577) -- (.147,.462);
  }\par\vspace{2pt}
  \textbf{Top view}\par\vspace{2pt}
  \checkview{\linewidth}{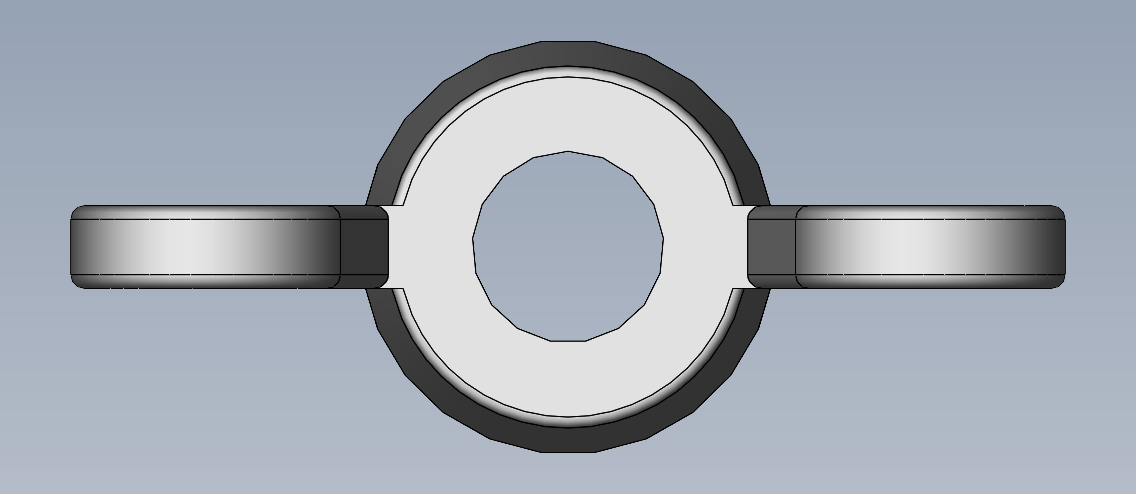}{%
    \node[checkbadge,fill=checkblue] at (.317,.198) {A};
    \draw[checkline,checkblue] (.357,.259) -- (.43,.381);
    \node[checklabel,font=\footnotesize,text=checkblue,inner sep=1pt]
      at (.479,.538) {$\varnothing$13.856};
    \draw[checkdim,checkteal] (.94,.416) -- (.94,.584);
    \draw[checkteal,line width=.45pt] (.89,.416) -- (.967,.416)
                                       (.89,.584) -- (.967,.584);
    \node[checklabel,font=\footnotesize,text=checkteal,anchor=east,inner sep=1pt]
      at (.959,.79) {6\,mm};
    \node[checkbadge,fill=checkteal] at (.751,.779) {B};
    \draw[checkline,checkteal] (.802,.728) -- (.86,.582);
  }%
  \end{minipage}%
}

\newcommand{\checkflangeviews}{%
  \begin{tabular}{@{}c@{\hspace{.025\columnwidth}}c@{}}
  {\small\bfseries Initial part} & {\small\bfseries Reference edit} \\[2pt]
  \checkview{.4875\columnwidth}{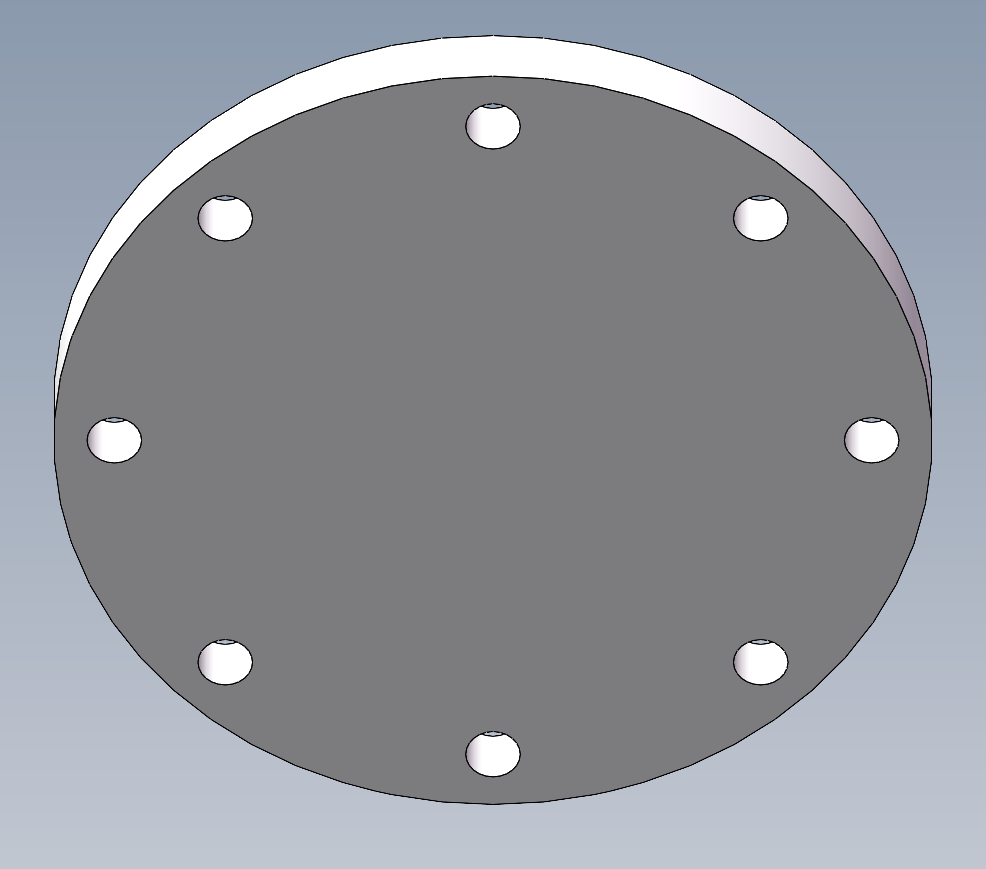}{%
    \node[checkbadge,fill=checkblue] at (.12,.63) {E};
    \draw[checkline,white] (.17,.618) -- (.44,.53);
  } &
  \checkview{.4875\columnwidth}{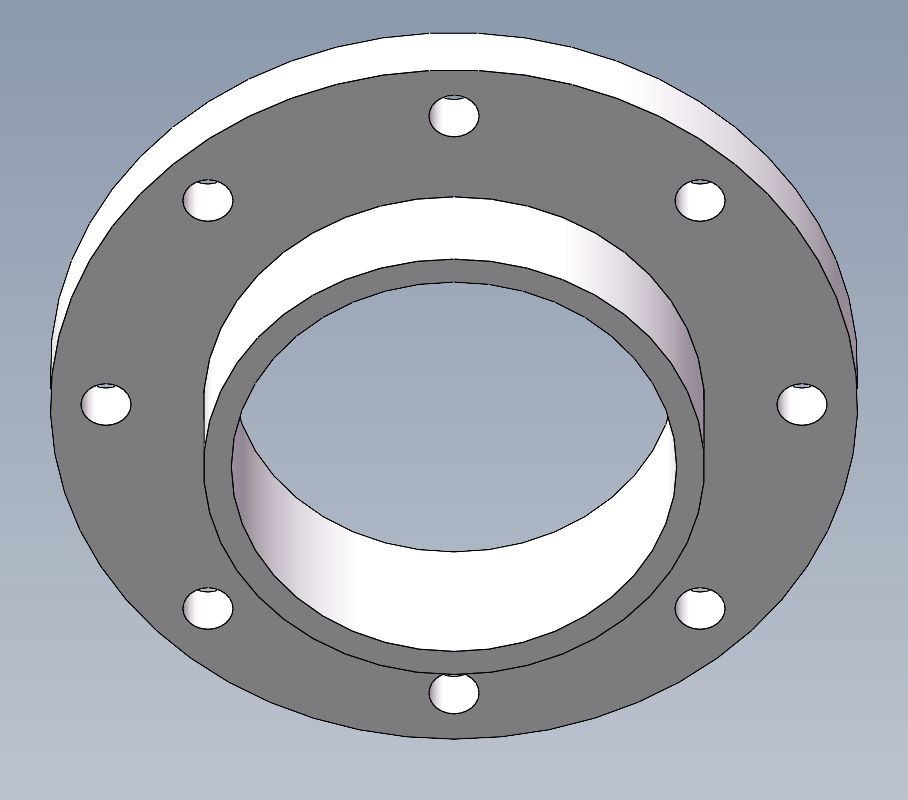}{%
    \node[checkbadge,fill=checkblue] at (.12,.63) {E};
    \draw[checkline,white] (.17,.618) -- (.46,.54);
  }\\
  \end{tabular}\\[4pt]
  \checkview{\columnwidth}{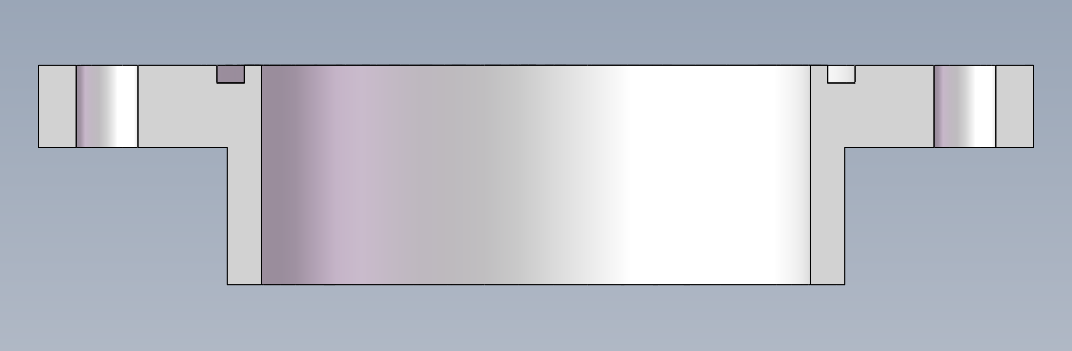}{%
    \draw[checkblue,densely dashed,line width=.55pt] (.5,.24) -- (.5,.795);
    \draw[checkdim,checkblue] (.244,.45) -- (.755,.45);
    \node[checklabel,text=checkblue] at (.5,.45) {$\varnothing$80};
    \draw[checkdim,checkrust] (.213,.908) -- (.787,.908);
    \draw[checkrust,line width=.45pt]
      (.213,.808) -- (.213,.945) (.787,.808) -- (.787,.945);
    \node[checklabel,text=checkrust] at (.5,.912) {$\varnothing$90};
    \draw[checkdim,checkrust] (.952,.425) -- (.952,.813);
    \draw[checkrust,line width=.45pt]
      (.801,.425) -- (.968,.425) (.803,.813) -- (.968,.813);
    \node[checklabel,text=checkrust,anchor=east] at (.93,.625) {20\,mm};
    \node[checkbadge,fill=checkrust] at (.075,.69) {D};
    \draw[checkline,checkrust] (.11,.68) -- (.213,.425);
    \draw[checkline,checkrust] (.11,.70) -- (.213,.813);
    \node[checkbadge,fill=checkteal] at (.125,.101) {F};
    \draw[checkline,checkteal] (.159,.13) -- (.213,.196);
  }%
}

%% file: sec/supplementary/11_evaluator_evidence.tex
\subsection{Evaluation Workflow}
\label{app:evaluator-evidence}

\begingroup
\setlength{\columnsep}{46pt}
\begin{multicols}{2}
\noindent Figure~\ref{fig:supp-evaluation-agent} shows how the agent
selects objects and measurements for checklist and process judgments.
Each judgment identifies the objects checked, explains the decision, and
cites the evidence. The scoring script validates the submission.

\noindent For the flanged rod with a side key, Claude Opus~5's Text2CAD output passed
17 of 18 conditions. Figure~\ref{fig:evaluator-case-views} locates
the checked regions, and
Table~\ref{tab:evaluator-full-checklist} links each verdict to its evidence.
\end{multicols}
\endgroup
\begingroup
\definecolor{evblue}{HTML}{214D6B}
\definecolor{evrust}{HTML}{9A422E}
\definecolor{evgreen}{HTML}{29604A}
\definecolor{evline}{HTML}{A9BFCB}
\definecolor{evfill}{HTML}{EFF4F7}
\definecolor{evsoftred}{HTML}{FAF0EC}
\definecolor{evcodebar}{HTML}{E2EAF0}
\definecolor{evgutter}{HTML}{E9EFF3}
\definecolor{evmuted}{HTML}{4E6272}
\definecolor{evpurple}{HTML}{67478D}
\newcommand{\evmonofont}{\fontencoding{T1}\fontfamily{zi4}\fontseries{m}\selectfont}

\begingroup
\setlength{\intextsep}{5pt}
\begin{figure}[H]
\centering
\begin{minipage}{.76\linewidth}
\centering
\begin{tikzpicture}[x=\linewidth,y=-.76pt,
  viewhead/.style={anchor=north west,inner sep=0pt,font=\footnotesize\bfseries,text=evblue},
  viewnote/.style={anchor=north,inner sep=0pt,font=\footnotesize,text=evmuted},
  viewport/.style={anchor=north west,inner sep=0pt},
  viewtag/.style={circle,fill=white,draw,line width=.55pt,minimum size=12pt,
    inner sep=0pt,font=\footnotesize\bfseries},
  viewleader/.style={line cap=round,line width=.65pt}]
  \useasboundingbox (0,0) rectangle (1,191);
  \fill[evfill] (0,0) rectangle (.551,190) (.569,0) rectangle (1,190);
  \fill[evcodebar] (0,0) rectangle (.551,18) (.569,0) rectangle (1,18);
  \node[viewhead] at (.012,1) {Isometric \enspace \texttt{E006}};
  \node[viewhead] at (.581,1) {Front \enspace \texttt{E007}};
  \node[viewport] at (.095,22)
    {\includegraphics[height=115.52pt]{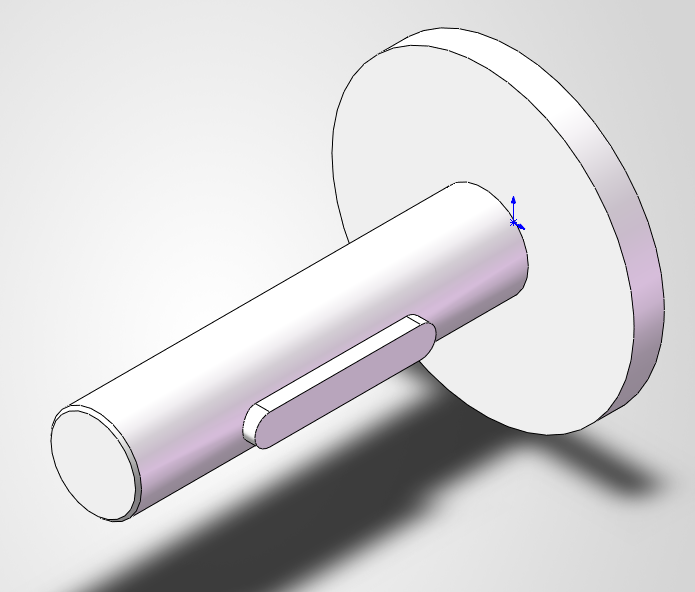}};
  \node[viewport] at (.630,22)
    {\includegraphics[height=115.52pt]{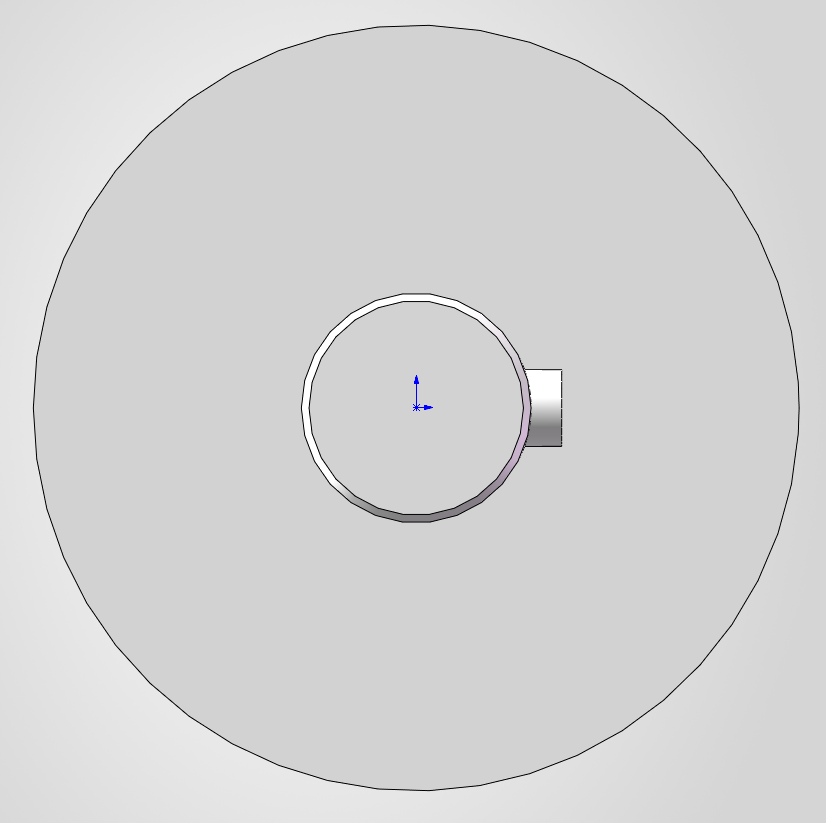}};

  \draw[viewleader,evblue] (.201,49) -- (.291,53);
  \draw[viewleader,evblue] (.201,49) -- (.216,105);
  \fill[evblue] (.291,53) circle[radius=1.25pt] (.216,105) circle[radius=1.25pt];
  \node[viewtag,draw=evblue,text=evblue] at (.201,49) {A};
  \draw[viewleader,evgreen] (.355,138) -- (.286,118);
  \fill[evgreen] (.286,118) circle[radius=1.25pt];
  \node[viewtag,draw=evgreen,text=evgreen] at (.355,138) {B};
  \draw[viewleader,evpurple] (.111,114) -- (.164,137);
  \fill[evpurple] (.164,137) circle[radius=1.25pt];
  \node[viewtag,draw=evpurple,text=evpurple] at (.111,114) {C};

  \node[viewnote] at (.2755,178) {Checklist groups A--C};
  \node[viewnote] at (.7845,178) {Coaxial flange and rod};
\end{tikzpicture}
\caption{SolidWorks views of the rebuilt flanged rod. Labels A--C locate the regions
checked in Table~\ref{tab:evaluator-full-checklist}.}
\label{fig:evaluator-case-views}
\end{minipage}
\end{figure}
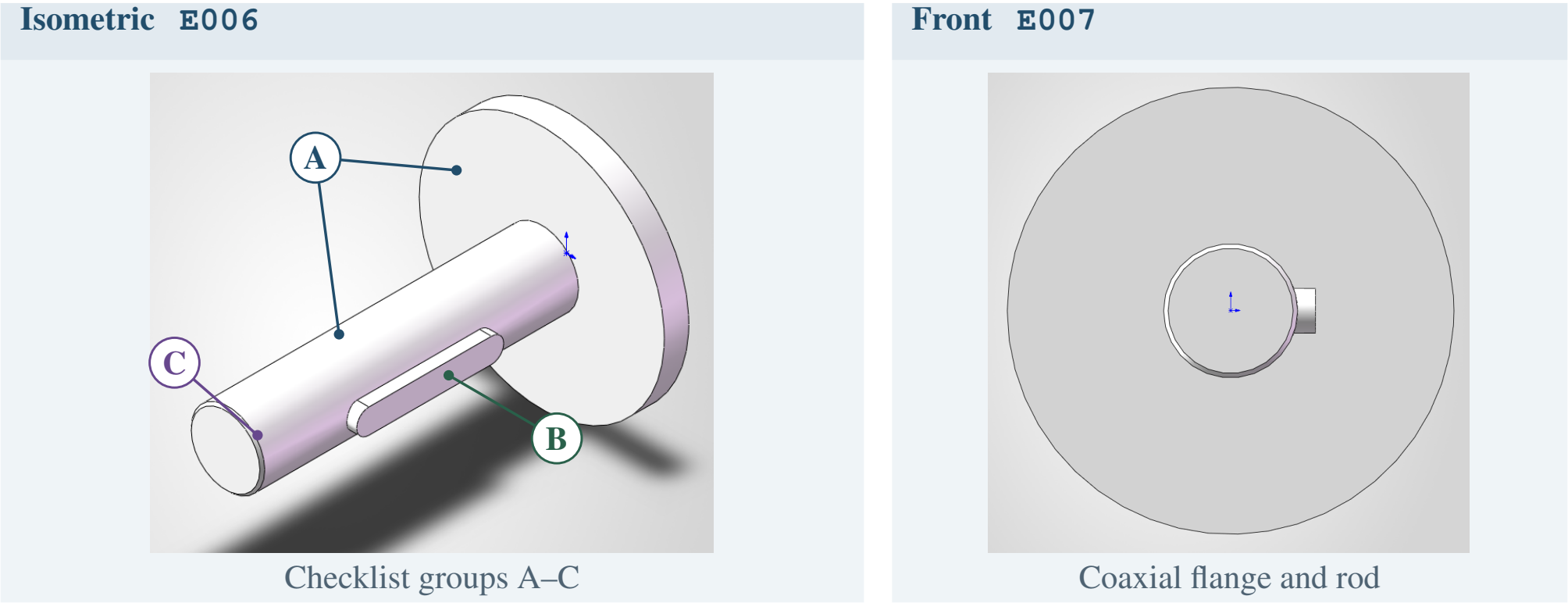

\begin{table}[H]
\centering
\caption{Checklist verdicts and evidence for the flanged rod generated from text.}
\label{tab:evaluator-full-checklist}
\small
\setlength{\tabcolsep}{5pt}
\renewcommand{\arraystretch}{1.08}
\begin{tabularx}{\linewidth}{l>{\raggedright\arraybackslash}Xll}
\toprule
ID & Acceptance condition & Verdict & Evidence IDs \\
\midrule
\multicolumn{4}{@{}l}{\textcolor{evblue}{\textbf{A}}\enspace\textit{Flange and rod}} \\
R01.01 & One connected solid comprising a flange, cylindrical rod, and side key. & Pass & E003, E004, E002 \\
R01.02 & Circular base flange. & Pass & E005, E006 \\
R01.03 & 50\,mm maximum outside diameter for the flange. & Pass & E005 \\
R01.04 & 5\,mm flange thickness. & Pass & E005, E004 \\
R01.05 & Upright cylindrical rod above the flange. & Pass & E005, E006 \\
R01.06 & 15\,mm rod diameter. & Pass & E005 \\
R01.07 & Coaxial rod and flange. & Pass & E005 \\
R01.08 & Rod axis perpendicular to the flange face. & Pass & E005 \\
R01.09 & 64.5\,mm from the top face of the flange to the rod tip. & Pass & E005, E004 \\
R01.10 & Overall part height of approximately 69.5\,mm. & Pass & E003, E004 \\
\addlinespace[2pt]
\multicolumn{4}{@{}l}{\textcolor{evgreen}{\textbf{B}}\enspace\textit{Side key}} \\
R02.01 & One side key with rounded ends. & Pass & E005, E006 \\
R02.02 & Continuous joint between the key and one side of the rod. & Pass & E005, E003 \\
R02.03 & 5\,mm key width. & Pass & E005 \\
\rowcolor{evsoftred}
\textbf{\hyperlink{a4-trace}{\textcolor{evrust}{R02.04}}} &
25\,mm overall key length along the rod axis. &
\textcolor{evrust}{\textbf{Fail}} &
\hyperlink{a4-e005}{E005}, \hyperlink{a4-e001}{E001}, \hyperlink{a4-e004}{E004} \\
R02.05 & 2.5\,mm radius at each end of the key. & Pass & E005 \\
R02.06 & Flat outer face of the key 9.5\,mm from the rod axis. & Pass & E005 \\
\addlinespace[2pt]
\multicolumn{4}{@{}l}{\textcolor{evpurple}{\textbf{C}}\enspace\textit{Chamfer and connectivity}} \\
R03.01 & 0.5\,mm $\times$ $45^\circ$ chamfer on the outer edge of the rod tip. & Pass & E005, E004 \\
R03.02 & Flange, rod, and key form a single connected solid. & Pass & E003, E004, E002 \\
\bottomrule
\end{tabularx}
\par\vspace{5pt}
\begin{minipage}{\linewidth}\raggedright
{\small\noindent
\textbf{Evidence IDs.} \texttt{E001} source code, \texttt{E002} run log,
\texttt{E003} build result, \texttt{E004} feature tree, \texttt{E005} analytic
faces, \texttt{E006} isometric view, and \texttt{E007} front view.
\par\smallskip
The side key passes the width and end-radius checks (R02.03 and R02.05).
Figure~\ref{fig:evaluator-trace} shows how the evaluator detects the
axial length violation (R02.04).
\par}
\end{minipage}
\end{table}
\endgroup

\newpage
\setlength{\intextsep}{3pt}

\begin{figure}[H]
\centering
\captionsetup{skip=4pt}
\definecolor{archtools}{HTML}{F4F3FB}
\definecolor{archagent}{HTML}{EBE8FB}
\definecolor{archjudgments}{HTML}{F1FBEF}
\definecolor{archscore}{HTML}{EFF4F7}
\begin{tikzpicture}[x=1pt,y=1pt,>=Stealth,
  module/.style={draw=black!36,rounded corners=2pt,line width=.45pt,
    align=center,font=\small,inner xsep=4pt,inner ysep=3pt,minimum height=36pt},
  flow/.style={-Stealth,draw=black!64,line width=.65pt},
  flow note/.style={font=\footnotesize,inner sep=1.5pt},
  retry/.style={flow,dashed,rounded corners=2pt}]
  \node[module,fill=archtools,text width=88pt] (criteria) at (43,0)
    {\textbf{Task criteria}\\Fixed checklist\\Six process dimensions};
  \node[module,fill=archagent,text width=96pt] (agent) at (153,0)
    {\textbf{Evaluation agent}\\Locate objects and\\select measurements};
  \node[module,fill=archtools,text width=115pt,minimum height=26pt]
    (tools) at (153,-44)
    {\textbf{Inspection tools}\\Code, logs, tree, faces, views};
  \node[module,fill=archjudgments,text width=132pt]
    (judgments) at (294,0)
    {\textbf{Judgment records}\\Checklist verdicts and evidence\\
     Process findings, links, and groups};
  \node[module,fill=archscore,text width=101pt] (score) at (431,0)
    {\textbf{Scoring script}\\Validate the record\\Compute scores};

  \draw[flow] (criteria) -- (agent);
  \draw[flow] (agent) -- (judgments);
  \draw[flow] (judgments) -- (score);
  \draw[flow] ([xshift=-21pt]agent.south) -- node[flow note,left] {query}
    ([xshift=-21pt]tools.north);
  \draw[flow] ([xshift=21pt]tools.north) -- node[flow note,right] {evidence}
    ([xshift=21pt]agent.south);
  \draw[retry] (score.south) -- (431,-44)
    -- node[flow note,above] {Validation errors} (220,-44)
    |- ([yshift=-10pt]agent.east);
\end{tikzpicture}
\caption{Evaluation agent architecture. Validation errors return to the
agent for resubmission. Accepted records are used to compute scores.}
\label{fig:supp-evaluation-agent}
\end{figure}
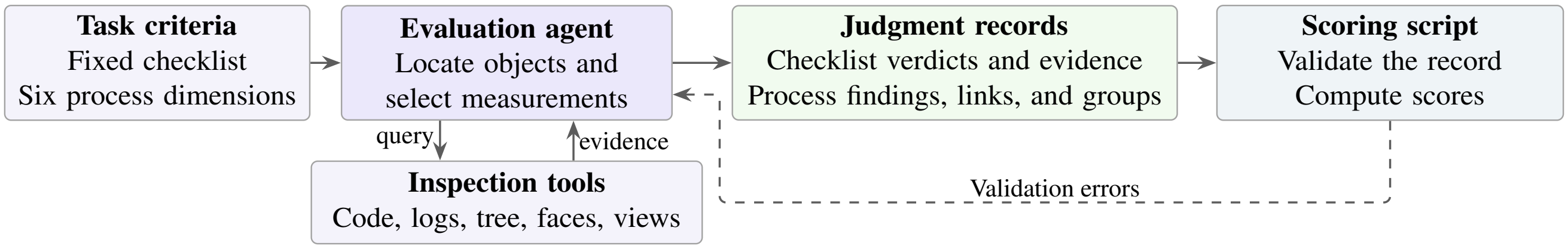

\noindent Figure~\ref{fig:evaluator-trace} shows that the submission passed
validation on Turn~6 with the verdict from Turn~3.
\par
\hypertarget{a4-trace}{}%

\begin{figure}[H]
\centering
\begin{tikzpicture}[x=.96\linewidth,y=-1pt,>=Stealth,
  trace text/.style={anchor=north west,inner sep=0pt,align=left,font=\normalsize},
  trace title/.style={trace text,font=\large\bfseries},
  trace label/.style={trace text,font=\normalsize\bfseries,text=evblue},
  trace code/.style={trace text,font=\fontsize{9.2}{10.2}\evmonofont},
  trace rule/.style={evline,line width=.45pt},
  trace step/.style={evblue,line width=.85pt,-Stealth},
  trace turn/.style={circle,draw=evblue,line width=.8pt,fill=white,text=evblue,
    minimum size=17pt,inner sep=0pt,font=\normalsize\bfseries},
  trace dimension/.style={Stealth-Stealth,line width=.9pt},
  trace measure/.style={font=\normalsize,inner sep=1.5pt,fill=white}]
  \useasboundingbox (0,0) rectangle (1.039,475);

  \newcommand{\evtracepanel}[2]{%
    \begin{scope}[shift={(0,#1)}]
      \begin{scope}
        \clip[rounded corners=2.5pt] (.069,0) rectangle (1.015,#2);
        \fill[evfill] (.069,0) rectangle (1.015,19);
      \end{scope}
      \draw[evline,line width=.6pt,rounded corners=2.5pt]
        (.069,0) rectangle (1.015,#2);
      \draw[trace rule] (.069,19) -- (1.015,19);
    \end{scope}%
  }

  \fill[evcodebar,rounded corners=2.5pt] (0,0) rectangle (1.015,22);
  \node[trace label] at (.014,6) {R02.04};
  \node[trace title] at (.102,4) {Side-key length along the rod axis};
  \node[trace title,text=evblue,anchor=north east] at (1,4) {25\,mm required};

  \evtracepanel{26}{125}
  \evtracepanel{157}{159}
  \evtracepanel{322}{99}
  \draw[trace step] (.028,44) -- (.028,157);
  \draw[trace step] (.028,175.5) -- (.028,322);
  \draw[trace rule] (.046,35.5) -- (.069,35.5)
                   (.046,166.5) -- (.069,166.5)
                   (.046,331.5) -- (.069,331.5);
  \node[trace turn] at (.028,35.5) {1};
  \node[trace turn] at (.028,166.5) {2};
  \node[trace turn,font=\small\bfseries] at (.028,331.5) {3--6};
  \node[trace title] at (.087,30) {Build records};
  \node[trace title] at (.087,161) {Native measurements};
  \node[trace title] at (.087,326) {Verdict and validation};

  \node[trace label] at (.087,51) {E001 \enspace PySW source};
  \fill[evfill] (.087,65) rectangle (.525,146);
  \draw[trace rule] (.141,65) -- (.141,146);
  \node[trace code,text=evmuted,anchor=north east,align=right] at (.130,65)
    {30\\\strut\\98\\99\\100\\101\\102\\103};
  \node[trace code] at (.151,65) {%
    key\_total\_length = \textcolor{evpurple}{25} * mm\\
    \textcolor{evmuted}{...}\\
    key\_sketch.\textcolor{evblue}{StraightSlot}(\\
    \hspace{2em}(key\_bottom\_z, \textcolor{evpurple}{0}),\\
    \hspace{2em}(key\_top\_z, \textcolor{evpurple}{0}),\\
    \hspace{2em}key\_width,\\
    \hspace{2em}full\_length=\textcolor{evpurple}{True},\\
    )};

  \node[trace label] at (.551,51) {E002 \enspace SolidWorks run log};
  \node[trace code,text=evmuted] at (.551,67) {25};
  \node[trace code] at (.608,67) {[side\_key\_extrusion]\\
    \textcolor{evgreen}{Feature created successfully}};
  \draw[trace rule] (.551,96) -- (1,96);
  \node[trace label] at (.551,104) {E003};
  \node[trace code] at (.628,104)
    {"force\_rebuild\_ok": \textcolor{evpurple}{true}};
  \node[trace label] at (.551,123) {E004};
  \node[trace code] at (.628,123)
    {"name": \textcolor{evblue}{"side\_key\_extrusion"}};

  \begin{scope}[shift={(0,10)}]
    \node[trace label] at (.087,174) {E006 \enspace Isometric view};
    \node[trace label] at (.391,174) {E005 \enspace Faces 6 and 8};
    \node[trace text,text=evmuted,anchor=north east] at (1,174) {SolidWorks right view};
    \node[anchor=north west,inner sep=0pt] at (.087,188)
      {\includegraphics[width=.253\linewidth]{figures/evaluator-c141-isometric.png}};
    \node[trace measure,text=evblue] at (.136,232) {Side key};
    \draw[trace step] (.153,236) -- (.217,255);
    \draw[trace step] (.356,242) -- (.384,242);
    \node[anchor=north west,inner sep=0pt] at (.391,187)
      {\includegraphics[width=.58464\linewidth]{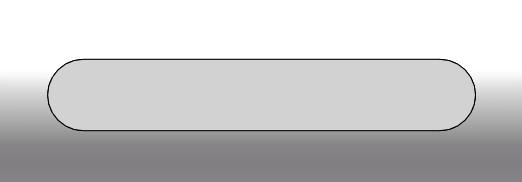}};
    \draw[evblue,line width=.62pt,densely dashed] (.488,203) -- (.488,240)
                                               (.903,203) -- (.903,240);
    \fill[evblue] (.488,240) circle[radius=1.4pt] (.903,240) circle[radius=1.4pt];
    \draw[evblue,trace dimension] (.488,202) -- (.903,202);
    \node[trace measure,text=evblue] at (.6955,202)
      {\textbf{25\,mm} between arc centers};
    \draw[evrust,line width=.62pt] (.446,240) -- (.446,273)
                                 (.945,240) -- (.945,273);
    \draw[evrust,trace dimension] (.446,273) -- (.945,273);
    \node[trace measure,text=evrust] at (.6955,273)
      {\textbf{30\,mm} end to end};
    \node[trace text] at (.391,292)
      {$z_{\max}-z_{\min}=(0.05225-0.02225)\,\mathrm{m}$};
  \end{scope}

  \begin{scope}[shift={(0,14)}]
    \node[trace label] at (.087,333) {Axial length};
    \node[trace text] at (.087,348) {Required};
    \node[trace text,anchor=north east] at (.514,348) {25\,mm};
    \node[trace text] at (.087,360) {Measured};
    \node[trace text,anchor=north east] at (.514,360) {30\,mm};
    \draw[trace rule] (.087,372) -- (.525,372);
    \node[trace text] at (.087,375) {Deviation};
    \node[trace text,anchor=north east] at (.514,375) {$+5$\,mm};

    \node[trace label] at (.551,333) {Turn 6 \enspace Accepted verdict};
    \fill[evfill] (.551,347) rectangle (1,383);
    \node[trace code] at (.560,351) {%
      "status": \textcolor{evrust}{"FAIL"},\\
      "decision\_basis": "observed\_violation",\\
      "evidence\_ids": ["E005", "E001", "E004"]};

  \node[trace text,font=\small] at (.087,391) {\textbf{3}\enspace Inconsistent score};
    \node[trace text,font=\small] at (.316,391) {\textbf{4}\enspace JSON parse error};
    \node[trace text,font=\small] at (.553,391) {\textbf{5}\enspace Missing fields};
    \node[trace text,font=\small\bfseries,text=evgreen] at (.793,391) {6 \enspace Accepted};
    \draw[trace step] (.278,397) -- (.304,397);
    \draw[trace step] (.515,397) -- (.541,397);
    \draw[trace step] (.751,397) -- (.778,397);
  \end{scope}

  \path[fill=evfill,draw=evline,line width=.6pt,rounded corners=2.5pt]
    (.069,427) rectangle (1.015,474);
  \node[trace label] at (.087,432) {Process attribution};
  \node[trace text,font=\small,text=evmuted,anchor=north east] at (1,433)
    {Shared evidence};
  \draw[trace rule] (.087,445) -- (1,445);
  \draw[trace rule] (.535,450) -- (.535,470);
  \node[trace text,font=\normalsize\bfseries,text=evrust] at (.087,449)
    {Dimensions fail};
  \node[trace text,font=\small,anchor=north east] at (.519,450)
    {\hyperlink{a4-e001}{E001}, \hyperlink{a4-e005}{E005}};
  \node[trace text,font=\small] at (.087,462) {25\,mm used as center spacing};
  \node[trace text,font=\normalsize\bfseries,text=evgreen] at (.551,449)
    {Sketch and combination pass};
  \node[trace text,font=\small,anchor=north east] at (1,450)
    {\hyperlink{a4-records}{E003--E006}};
  \node[trace text,font=\small] at (.551,462) {Rounded, joined key};

  \fill[evblue!76] (1.015,307) circle[radius=1.4pt];
  \draw[trace step,evblue!76,densely dashed,rounded corners=2pt]
    (1.015,307) -- (1.038,307) -- (1.038,437) -- (1.015,437);
\end{tikzpicture}
\caption{Side-key length error. The sketch uses the required
25\,mm length as the arc-center spacing. Two 2.5\,mm end radii increase
the overall length to 30\,mm.\
\protect\hyperlink{a4-records}{See evidence excerpts.}}
\label{fig:evaluator-trace}
\end{figure}

\newpage

\begingroup
\setlength{\columnsep}{.024\linewidth}
\setlength{\multicolsep}{0pt}
\begin{multicols}{2}
\noindent\hypertarget{a4-records}{\textbf{Side-key length verification.}}
The feature tree (E004) identifies the side-key extrusion. The axial
bounds of rounded faces 6 and 8 (E005) span 30\,mm, and the source
(E001) places the arc centers 25\,mm apart.
\par\columnbreak
\noindent The R02.04 verdict cites
all three records. The five-field process record for this example attributes the error
to dimensions and parameters. Section~\ref{app:process-validation}
uses the six-dimension rubric.
\end{multicols}
\endgroup

\par\smallskip

\begingroup
\renewcommand{\ttdefault}{zi4}
\newcommand{\evturnhead}[2]{%
  \noindent{\small\bfseries\color{evblue}\strut Turn~#1\enspace #2\par}%
  \kern .6pt
  {\color{evline}\hrule height .45pt}}
\newcommand{\evrecordhead}[2]{%
  \par\vspace{4pt}\noindent
  \colorbox{evcodebar}{\parbox{\dimexpr\linewidth-2\fboxsep\relax}{%
    \footnotesize\bfseries\color{evblue}#1\hfill
    \normalfont\color{evmuted}#2}}\par\nobreak}
\lstdefinelanguage{EvaluatorJSON}{
  morestring=[b]",morekeywords={true,false,null},sensitive=true}
\lstset{basicstyle=\fontsize{8.4}{9.3}\evmonofont,columns=fullflexible,
  keepspaces=true,backgroundcolor=\color{evfill},frame=none,
  xleftmargin=5pt,xrightmargin=5pt,framexleftmargin=5pt,framexrightmargin=5pt,
  aboveskip=0pt,belowskip=0pt,showstringspaces=false,
  breaklines=true,breakatwhitespace=true,breakindent=8pt,
  keywordstyle=\color{evpurple},stringstyle=\color{evblue},
  numberstyle=\fontsize{8}{9.3}\evmonofont\color{evmuted},numbersep=8pt,
  literate=*{0}{{\textcolor{evpurple}{0}}}1
    {1}{{\textcolor{evpurple}{1}}}1 {2}{{\textcolor{evpurple}{2}}}1
    {3}{{\textcolor{evpurple}{3}}}1 {4}{{\textcolor{evpurple}{4}}}1
    {5}{{\textcolor{evpurple}{5}}}1 {6}{{\textcolor{evpurple}{6}}}1
    {7}{{\textcolor{evpurple}{7}}}1 {8}{{\textcolor{evpurple}{8}}}1
    {9}{{\textcolor{evpurple}{9}}}1}
\lstdefinestyle{evnumbered}{numbers=left,xleftmargin=24pt,framexleftmargin=24pt}
\lstdefinestyle{evpython}{style=evnumbered,language=Python,
  morekeywords={True,False,None},emph={StraightSlot},emphstyle=\color{evblue}}
\begingroup
\noindent
\begin{minipage}[t]{.488\linewidth}
\evturnhead{1}{Code and build records}
\evrecordhead{Generator model}{JSON}
\begin{lstlisting}[language=EvaluatorJSON]
"model": "anthropic/claude-opus-5"
\end{lstlisting}
\evrecordhead{\hypertarget{a4-e001}{E001} \enspace PySW source}{Python}
\begin{lstlisting}[style=evpython,firstnumber=30]
key_total_length = 25 * mm
\end{lstlisting}
\begin{lstlisting}[style=evnumbered,numbers=none,basicstyle=\fontsize{8.4}{9.3}\evmonofont\color{evmuted}]
...
\end{lstlisting}
\begin{lstlisting}[style=evpython,firstnumber=98]
key_sketch.StraightSlot(
    (key_bottom_z, 0),
    (key_top_z, 0),
    key_width,
    full_length=True,
)
\end{lstlisting}
\evrecordhead{E002 \enspace Execution log}{Log}
\begin{lstlisting}[style=evnumbered,firstnumber=25,
  emph={successfully},emphstyle=\color{evgreen}]
[side_key_extrusion] Feature created successfully
\end{lstlisting}
\evrecordhead{E003 \enspace Build result}{JSON}
\begin{lstlisting}[language=EvaluatorJSON]
"status": "executed",
"body_count": 1,
"force_rebuild_ok": true
\end{lstlisting}
\evrecordhead{\hypertarget{a4-e004}{E004} \enspace Native feature tree}{JSON}
\begin{lstlisting}[language=EvaluatorJSON]
"tree_index": 24,
"name": "side_key_extrusion",
"type": "Boss"
\end{lstlisting}
\end{minipage}\hfill
\begin{minipage}[t]{.488\linewidth}
\evturnhead{2}{Native geometry}
\evrecordhead{\hypertarget{a4-e005}{E005} \enspace Analytic face records}{JSON}
\begin{lstlisting}[language=EvaluatorJSON]
"face_index": 6,
"origin_m": [0.0095, 0.0, 0.024749999999999998],
"radius_m": 0.0025,
"bbox_m": [
  0.006918124148940556, -0.0025, 0.02225,
  0.0095, 0.0025, 0.024749999999999998
]
...
"face_index": 8,
"origin_m": [0.0095, 0.0, 0.04975],
"radius_m": 0.0025,
"bbox_m": [
  0.006918124148940555, -0.0025, 0.04975,
  0.0095, 0.0025, 0.052250000000000005
]
\end{lstlisting}
\evrecordhead{Turn 6 \enspace Accepted verdict}{JSON}
\begin{lstlisting}[language=EvaluatorJSON]
"id": "R02.04",
"status": "FAIL",
"attempt": "attempted",
"realization": "executed_binding",
"evidence_ids": ["E005", "E001", "E004"],
"decision_basis": "observed_violation"
\end{lstlisting}
\end{minipage}
\par\smallskip
\endgroup
\endgroup

{\small\noindent In this example, the evaluator applies tolerances of 0.01\,mm to nominal lengths,
$0.1^\circ$ to nominal angles, and 3\% to approximate envelopes. Its
decision basis distinguishes observed violations, incomplete coverage,
and insufficient evidence.\par}
\endgroup

%% file: sec/supplementary/12_baseline_adaptation.tex
\subsection{Requirement Evaluator Comparison}
\label{app:baseline-adaptation}
\begin{multicols}{2}
All evaluators in Table~6 of the main paper use the same acceptance
checklist per candidate across three trials. An undecided verdict reduces
coverage and counts as a miss when the reference indicates a violation.

\begin{figure}[H]
\centering
\begin{tikzpicture}[x=1pt,y=-1pt,>=Stealth,
  box/.style={rounded corners=2pt,draw=black!33,line width=.45pt,
    align=center,font=\footnotesize,text width=96pt,inner xsep=3pt,inner ysep=3pt},
  evidence/.style={box,fill=blue!4},
  decision/.style={box,fill=green!6},
  native/.style={decision,fill=blue!6},
  method title/.style={font=\small\bfseries,inner sep=0pt},
  flow/.style={-Stealth,draw=black!62,line width=.55pt},
  output/.style={font=\footnotesize,inner sep=2pt}]
  \node[method title] at (60,0) {(a) BlenderLLM};
  \node[method title] at (181,0) {(b) MUSE};
  \node[method title] at (302,0) {(c) CADTests};
  \node[method title] at (428,0) {(d) PySWBench};

  \node[evidence] (blviews) at (60,24) {PySW script and\\four rendered views};
  \node[evidence] (blcriteria) at (60,61) {Checklist};
  \node[decision] (bljudge) at (60,116) {Multimodal judge};
  \draw[flow] (blviews.west) -- ++(-7pt,0) |- (bljudge.west);
  \draw[flow] (blcriteria) -- (bljudge);

  \node[evidence] (mref) at (181,24) {Checklist and\\reference drawings};
  \node[decision] (mrubric) at (181,55) {Derive a part rubric};
  \node[evidence] (mviews) at (181,83) {Reference and\\candidate views};
  \node[decision] (mjudge) at (181,116) {Compare views\\against the rubric};
  \draw[flow] (mref) -- (mrubric);
  \draw[flow] (mrubric.west) -- ++(-7pt,0) |- (mjudge.west);
  \draw[flow] (mviews) -- (mjudge);

  \node[evidence] (cref) at (302,24) {Checklist and\\reference solid};
  \node[decision] (ctests) at (302,55) {Generate tests};
  \node[decision] (crefine) at (302,83) {Run on reference\\and modified solids};
  \node[decision] (cjudge) at (302,116) {Run fixed tests on\\candidate STEP};
  \draw[flow] (cref) -- (ctests);
  \draw[flow] (ctests) -- (crefine);
  \draw[flow] (crefine.east) -- ++(7pt,0) |- (ctests.east);
  \node[font=\scriptsize,rotate=90,inner sep=0pt] at (366,69) {refine};
  \draw[flow] (crefine) -- (cjudge);

  \node[evidence] (pinput) at (428,24) {Native part, PySW script,\\and build logs};
  \node[native] (pagent) at (428,55) {Agent selects objects\\and measurements};
  \node[evidence] (pquery) at (428,85) {Query geometry, feature\\tree, and views};
  \node[decision] (pjudge) at (428,116) {Judge checklist conditions};
  \draw[flow] (pinput) -- (pagent);
  \draw[flow] ([xshift=-13pt]pagent.south) -- ([xshift=-13pt]pquery.north);
  \draw[flow] ([xshift=13pt]pquery.north) -- ([xshift=13pt]pagent.south);
  \draw[flow] (pquery) -- (pjudge);

  \foreach \x/\judge in {60/bljudge,181/mjudge,302/cjudge}{
    \node[output] (verdicts) at (\x,144) {Checklist verdicts};
    \draw[flow] (\judge) -- (verdicts);
  }
  \node[output] (pverdicts) at (428,144) {Verdicts with evidence IDs};
  \draw[flow] (pjudge) -- (pverdicts);
\end{tikzpicture}
\caption{Requirement evaluation in SolidWorks. PySWBench selects
candidate-specific measurements and cites evidence for each verdict.}
\label{fig:supp-baseline-adaptation}
\end{figure}
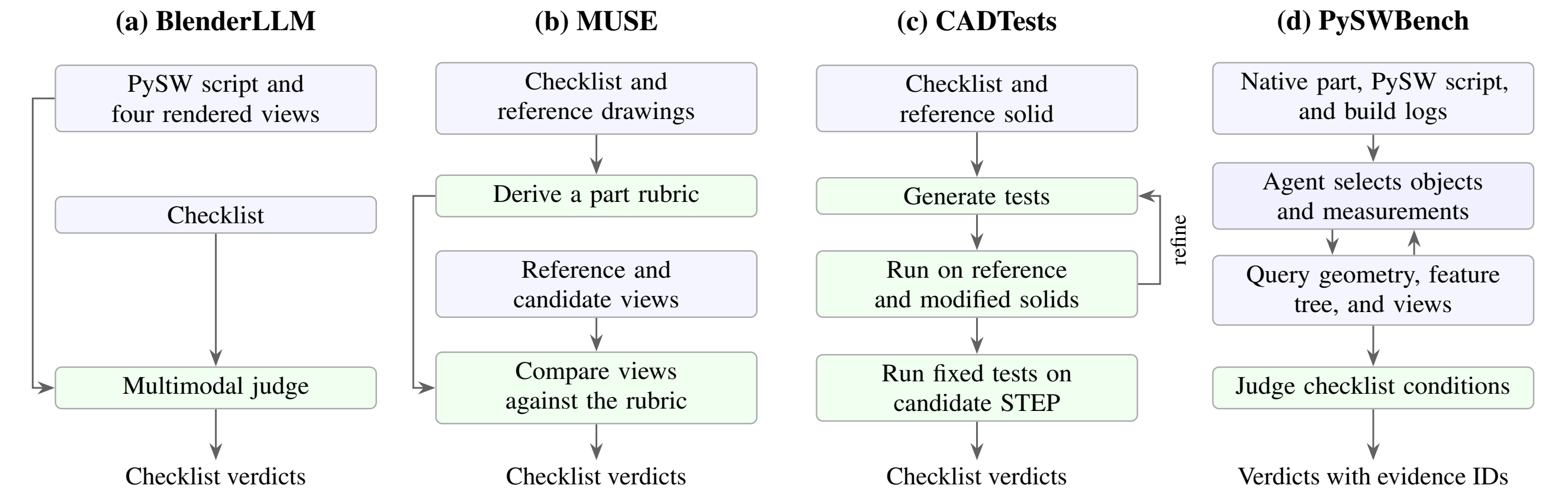

Figure~\ref{fig:supp-baseline-adaptation} compares the four evaluators.
MUSE renders views with hidden edges in CadQuery and reuses a rubric
derived from the reference drawings. CADTests refines tests on
transformed and edited reference solids. Its read-only fallback uses
saved SolidWorks measurements for supported queries.
Figure~\ref{fig:supp-evaluation-agent} details the PySWBench inspection loop.
\end{multicols}
\clearpage

%% file: sec/supplementary/13_process_validation.tex
\subsection{Process Validation}
\label{app:process-validation}
\begin{multicols}{2}
We randomly sampled 84 outputs from seven models, 42 per modality
(Text2CAD and Drawing2CAD), and evaluated them in three independent
runs with fixed inputs and settings. For each output and dimension,
the reference label is pass if evidence confirms all applicable checks,
fail if it establishes an error, and unresolved otherwise.

For 467 output--dimension pairs, the requirement evaluation provides enough
evidence for experts to determine whether the dimension passes or fails.
The process evaluator agrees with their judgments in 90.36\% of these cases
on average across runs. Counting the remaining 37 inconclusive cases as
having no confirmed error gives 88.49\% agreement on error status across
all 504 pairs (Table~\ref{tab:supp-process-validation}).

Across 84 modeling traces, we found evidence of error propagation in
20 outputs and confirmed 35 propagation links across those outputs.
A link counts as recovered when the evaluator identifies the same upstream
and downstream dimensions in the same output. Mean recall across the
three runs was 79.05\%.
\end{multicols}

\begin{table}[H]
\centering
\caption{Agreement on six process dimensions for 84 generation outputs.
Rates are percentages, reported as means and
sample standard deviations across three independent runs.}
\label{tab:supp-process-validation}
\small
\begin{tabular}{@{}lrrr@{}}
\toprule
Dimension & Definite labels & Accuracy on definite labels & Error agreement on all outputs\\
\midrule
Intent coverage    & 74 & $88.29 \pm 1.56$ & $88.10 \pm 2.06$\\
Modeling plan      & 84 & $93.25 \pm 1.82$ & $93.25 \pm 1.82$\\
Sketch contour     & 79 & $85.23 \pm 1.93$ & $83.73 \pm 1.82$\\
Sketch parameters  & 76 & $96.05 \pm 1.32$ & $92.86 \pm 1.19$\\
Spatial placement  & 75 & $94.22 \pm 3.36$ & $88.89 \pm 2.75$\\
Feature construction & 79 & $85.23 \pm 4.45$ & $84.13 \pm 3.64$\\
\midrule
Overall & 467 & $90.36 \pm 1.34$ & $88.49 \pm 1.37$\\
\bottomrule
\end{tabular}
\par\smallskip
{\footnotesize Directional recall on 35 reference links in 20 outputs: $(79.05 \pm 7.19)\%$.}
\end{table}

%% file: sec/supplementary/14_annotation_records.tex
\subsection{Reference Annotation}
\label{app:annotation-records}
\begin{multicols}{2}
\noindent A CAD expert assigned the requirement labels using evidence
from source code, execution records, and rebuilt geometry. GPT-5.6-Sol
in Codex helped locate the evidence. The expert inspected native
SolidWorks parts as needed and left inconclusive checks unresolved.
We finalized the labels before comparing them with evaluator predictions.
Figure~\ref{fig:supp-blind-review}(a) illustrates how the expert assigns a requirement label.
Panels (b) and (c) illustrate process annotation.
\end{multicols}
\begingroup
\setlength{\intextsep}{5pt}
\captionsetup{skip=5pt}
\newcommand{\recordhead}[1]{\par\smallskip\noindent{\color{blue!35!black}\bfseries #1}\par}
\newcommand{\recordfile}[2]{\hyperlink{blind-#1}{\texttt{#2}}}
\newcommand{\recordtarget}[2]{\hypertarget{blind-#1}{\texttt{#2}}}
\begin{figure}[H]
\centering
\begin{minipage}{.96\linewidth}
\small
\textbf{(a) Through-hole verification \hfill Text2CAD / R03.03}\par
\textit{The central hole must pass through both hexagonal bosses and the middle body.}\par
\recordfile{cut}{part-A/code.py}\quad
\recordfile{log}{part-A/log.log}\quad
\recordfile{bore}{part-A/native.json}\par\smallskip
\begin{minipage}[t]{.43\linewidth}\vspace{0pt}\centering
\includegraphics[width=.48\linewidth,height=73pt,keepaspectratio]{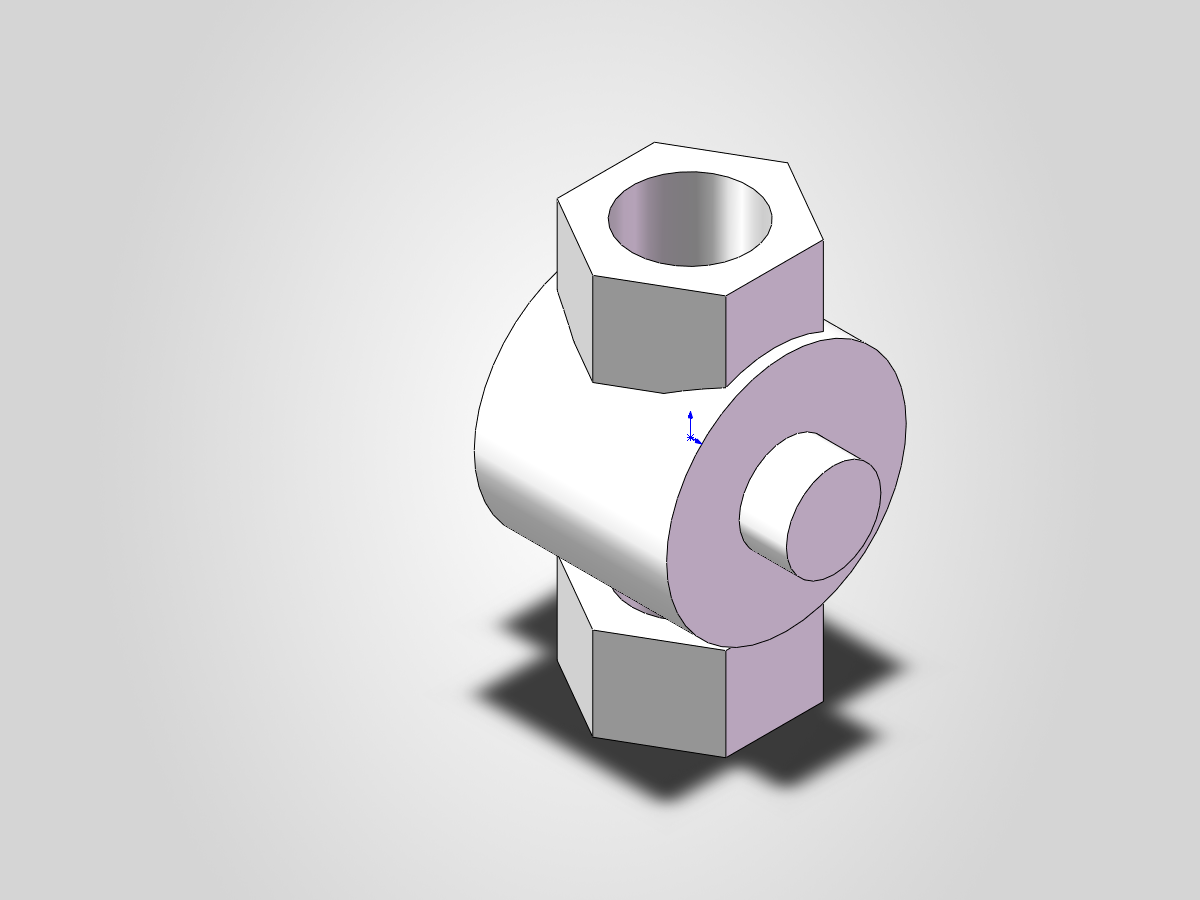}\hfill
\includegraphics[width=.48\linewidth,height=73pt,keepaspectratio]{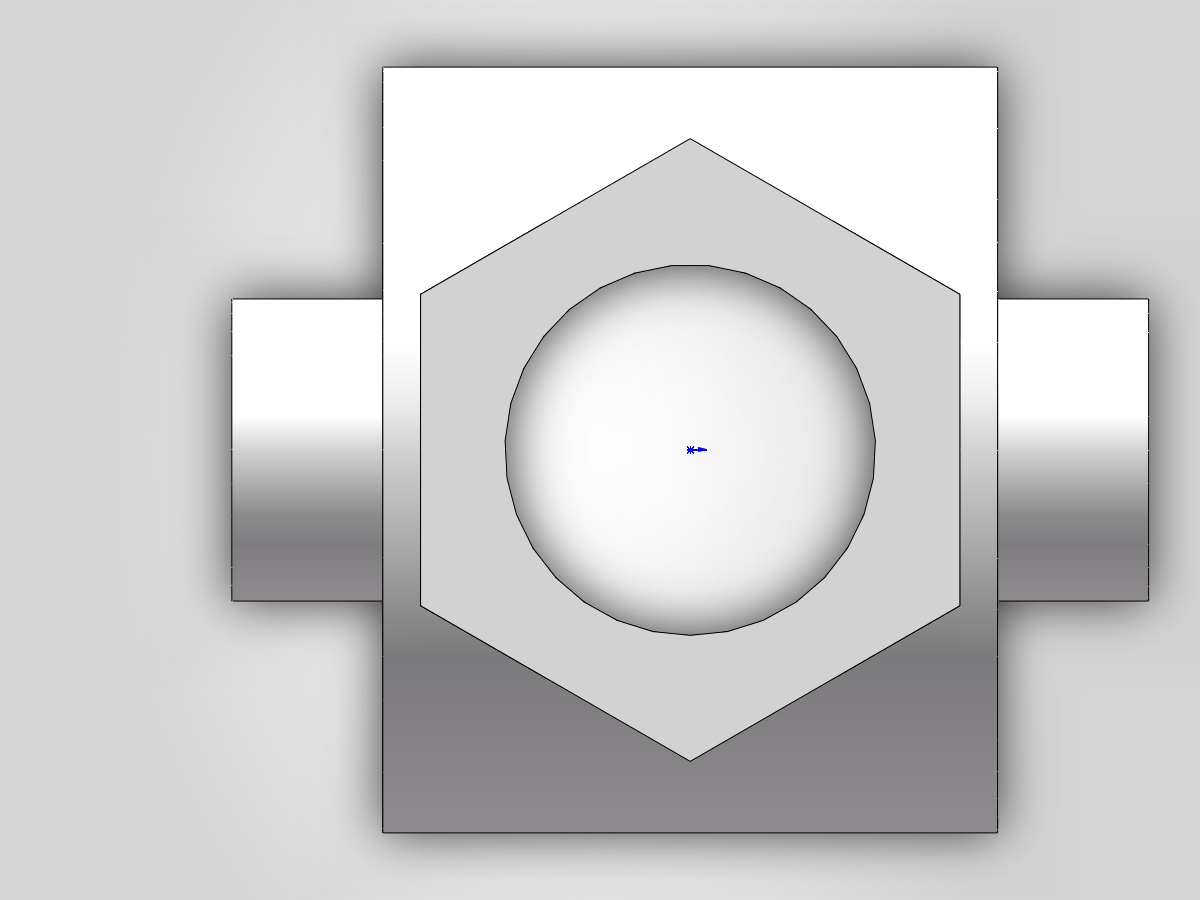}\\
{\footnotesize Isometric and top views of the rebuilt part}
\end{minipage}\hfill
\begin{minipage}[t]{.54\linewidth}\vspace{0pt}
\recordhead{Cut definition}
\recordtarget{cut}{code.py} (lines 143, 146--147)\par
{\ttfamily\footnotesize central\_hole = part.Features.ExtrudeCut(\\
\hspace*{1em}\ldots\\
\hspace*{1em}end\_condition="through\_all\_both",\\
\hspace*{1em}feature\_name="central\_through\_hole",}
\end{minipage}
\par\noindent \recordtarget{log}{log.log}, line~30, confirms the cut.
In \recordtarget{bore}{native.json}, cylindrical face \texttt{b0:f25}
spans the two end planes along the common boss axis.
\par\noindent\textbf{Expert verdict.} Pass. The bore opens through both
bosses and the middle body.
\par\noindent\textbf{Native model.} \texttt{review-models/part-A/output.SLDPRT}.
Section the bore along its axis to inspect both openings.
\end{minipage}
\caption{Reference annotation examples. (a) The expert verifies that the
bore passes through both bosses and the middle body.}
\label{fig:supp-blind-review}
\end{figure}

\begin{figure}[H]
\ContinuedFloat
\captionsetup{list=no}
\centering
\begin{minipage}{.96\linewidth}
\small
\textbf{(b) Missing rollers \hfill Drawing2CAD / Feature}\par
\textit{The chain requires two rows of rollers, with each roller coaxial with its pin.}\par
\recordfile{pin}{part-B/source.py}\quad
\recordfile{chain}{part-B/native.json}\quad
\recordfile{notes}{part-B/reviews/A,B}\par\smallskip
\begin{minipage}[t]{.36\linewidth}\vspace{0pt}\centering
\includegraphics[width=\linewidth,height=82pt,keepaspectratio]{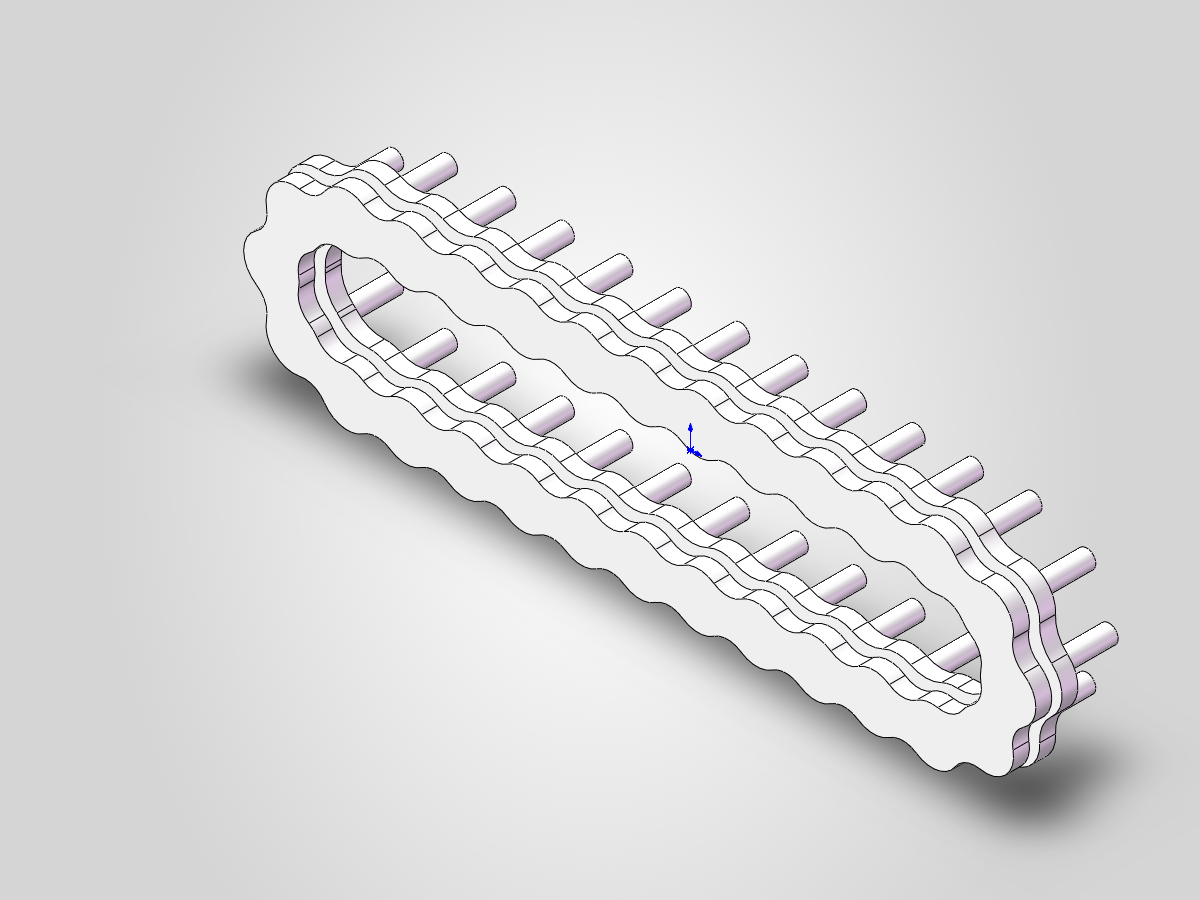}\\
{\footnotesize Rebuilt chain candidate}
\end{minipage}\hfill
\begin{minipage}[t]{.61\linewidth}\vspace{0pt}
\recordhead{Pin construction}
\recordtarget{pin}{source.py}, E004, lines 196--201, sketches and extrudes
the pins as one set.\par
{\ttfamily\footnotesize sk\_pins.Circle(points[i],\\
\hspace*{1em}diameter=pin\_diameter, \ldots)\\
\ldots\ feature\_name="pin\_set"}\par
The tree in \recordtarget{chain}{native.json} lists plates and pins.
The recorded cylindrical faces belong to the pins.
\end{minipage}
\recordhead{Independent reviews}
Of the two DeepSeek reviews in \recordtarget{notes}{reviews/A,B},
one traced the pin extrusion and found no roller solid. The other
found only pin surfaces in the cylinder catalogue.
\par\smallskip\textbf{Reference verdict.} Feature fails because the
roller bodies are missing. The pins occupy the required axes. All three
saved evaluator judgments agree.
\par\smallskip\textbf{Native model.} \texttt{review-models/part-B/output.SLDPRT}.
Inspect a pin along its axis for a separate coaxial roller.
\par\smallskip\hrule
\recordhead{(c) Intent-to-Param propagation}
In \texttt{part-C/source.py}, line 23 assigns the misread cover width
to \texttt{tongue\_w}. Lines 108 and 119 use $+\texttt{tongue\_w}/2$ and
$-\texttt{tongue\_w}/2$ to place the two trim boundaries, linking the
misread dimension to the incorrect trim parameters.
\end{minipage}
\caption{Reference annotation examples (continued). (b) Missing rollers.
(c) Intent-to-Param propagation.}
\end{figure}
\endgroup

%% file: sec/supplementary/11_freecad.tex
\subsection{Transfer to FreeCAD}
\label{app:freecad-reliability}
\begin{multicols}{2}
We repeated the PySWBench requirement evaluation in FreeCAD using the
same candidates, checklists, and human annotations as the SolidWorks
evaluation in Table~6 of the main paper.

We translated each candidate's sketches and modeling operations into a
program that builds a native FreeCAD document. The evaluator reads the
program, execution log, feature tree, and views, and queries native
geometry to verify through-holes, measure faces, and compare features
along a specified axis. Both platforms use the same evaluation prompt,
model settings, and tool-call budget.

The regular set contains 84 candidates and 1,642 conditions. The stress
set contains 23 candidates and 516 conditions. For each candidate, we
ran three independent sessions per platform and compared each verdict
with its human annotation.

\noindent Table~\ref{tab:freecad-reliability} compares the two platforms.
Both reached a verdict on every condition in all three trials.
\end{multicols}
\begin{table}[H]
\centering
\caption{Evaluator reliability in SolidWorks and FreeCAD on the regular
and stress sets. Entries are means and sample standard deviations across
three trials.}
\label{tab:freecad-reliability}
\small
\begin{tabular}{llrrrr}
\toprule
Set & Platform & Cohen's $\kappa$ & Acc (\%) & FNR (\%) & FPR (\%)\\
\midrule
Regular & SolidWorks & $0.850\pm0.032$ & $99.35\pm0.14$ & $20.51\pm2.56$ & $0.17\pm0.10$\\
        & FreeCAD & $0.830\pm0.017$ & $99.23\pm0.04$ & $17.95\pm6.78$ & $0.35\pm0.13$\\
\midrule
Stress  & SolidWorks & $0.762\pm0.005$ & $93.67\pm0.11$ & $31.21\pm2.21$ & $0.79\pm0.55$\\
        & FreeCAD & $0.732\pm0.041$ & $93.09\pm0.78$ & $35.82\pm6.41$ & $0.47\pm0.47$\\
\bottomrule
\end{tabular}
\end{table}
\begin{multicols}{2}
On FreeCAD, the evaluator's accuracy was only 0.12 percentage points lower
than on SolidWorks on the regular set and 0.58 points lower on the stress
set, showing that the requirement evaluation transfers to FreeCAD with
comparable per-condition accuracy. Building on this result, we will adapt
the full PySWBench benchmark to FreeCAD and release it alongside the
SolidWorks version. The release will include modeling and evaluation
prompts tailored to FreeCAD, a reproducible runtime environment, and
native inspection tools so that researchers without a SolidWorks license
can use PySWBench to evaluate the CAD modeling capabilities of large
language models.
\end{multicols}

%% file: sec/supplementary/15_complexity_trends.tex
\begin{multicols}{2}
\subsection{Performance Across Complexity Quintiles}
\label{app:complexity-trends}
Figure~\ref{fig:supp-complexity-trends} compares execution success and
requirement satisfaction (ReqSat) for seven models on the 150 parts grouped in
Section~\ref{app:dataset-complexity}. ReqSat assigns zero to tasks
without CAD output. On Text2CAD, mean execution success and mean ReqSat
decline with complexity and differ by at most two percentage points
in each quintile.

On Drawing2CAD, ReqSat trails execution success by about 19 percentage
points in Q1 and 25 in Q5. This gap is consistent with misread dimensions
and omitted features documented in the main paper.
\end{multicols}
\begingroup
\setlength{\intextsep}{4pt}
\captionsetup{skip=5pt}
\begin{figure}[H]
\centering
\includegraphics[width=.97\linewidth,trim=0 7bp 0 4bp,clip]{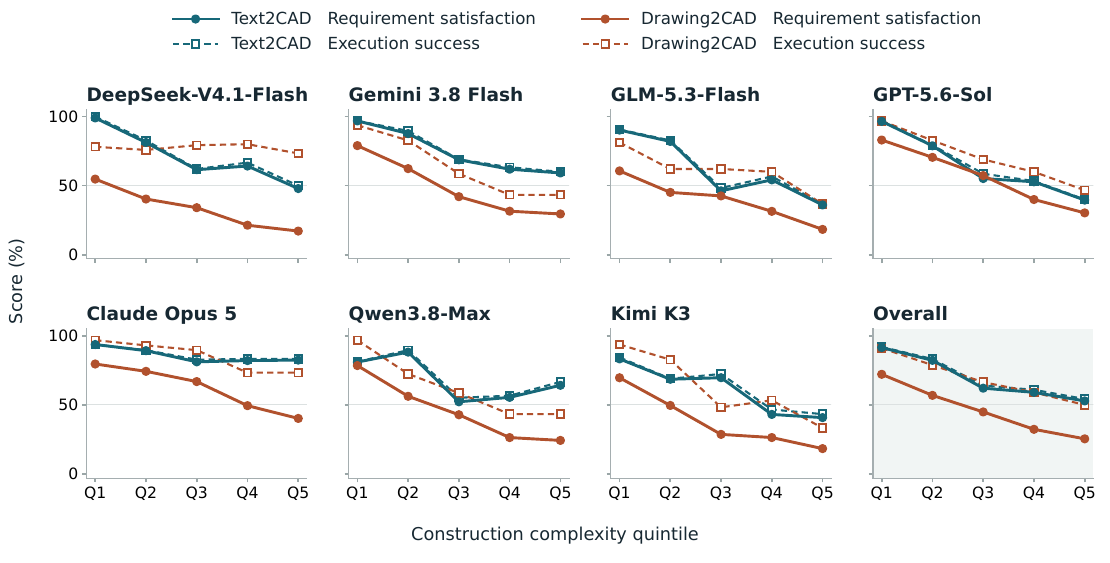}
\caption{Execution success and requirement satisfaction across construction
complexity quintiles from Q1 (lowest) to Q5 (highest). The Overall panel
averages the seven models within each quintile and input modality.}
\label{fig:supp-complexity-trends}
\end{figure}
\endgroup

%% file: sec/supplementary/16_complexity_factors.tex
\begin{multicols}{2}
\subsection{Construction Factors and ReqSat}
\label{app:complexity-factors}
Figure~\ref{fig:supp-factor-trends} groups the 150 parts by each
construction factor. Because the factors can covary,
Table~\ref{tab:supp-factor-correlations} reports each factor's partial
Spearman correlation with ReqSat after adjustment for the other two.

On Text2CAD, mean ReqSat falls with more sketch-plane changes, with a
weaker trend for maximum sketch size. The partial correlation for plane
changes has the largest magnitude among the three factors and remains modest.

On Drawing2CAD, ReqSat falls with both solid feature count and sketch-plane
changes. These counts are strongly correlated across the reference parts
(Spearman $\rho=0.87$). After adjustment, the correlation with plane
changes is close to zero. Solid feature count and maximum sketch size retain
negative correlations.

\end{multicols}

\begingroup
\setlength{\intextsep}{4pt}
\begin{table}[H]
\centering
\caption{Partial Spearman correlations between construction factors and
ReqSat. Each coefficient adjusts for the other two factors. We compute
correlations over 150 parts per model and average across seven models.}
\label{tab:supp-factor-correlations}
\begin{tabular}{lrr}
\toprule
Construction factor & Text2CAD & Drawing2CAD \\
\midrule
Solid features        & $-0.08$ & $-0.21$ \\
Sketch-plane changes  & $-0.14$ & $-0.04$ \\
Maximum sketch size   & $-0.04$ & $-0.19$ \\
\bottomrule
\end{tabular}
\end{table}

\begin{figure}[H]
\captionsetup{skip=5pt}
\centering
\includegraphics[width=.99\linewidth,trim=0 4bp 0 4bp,clip]{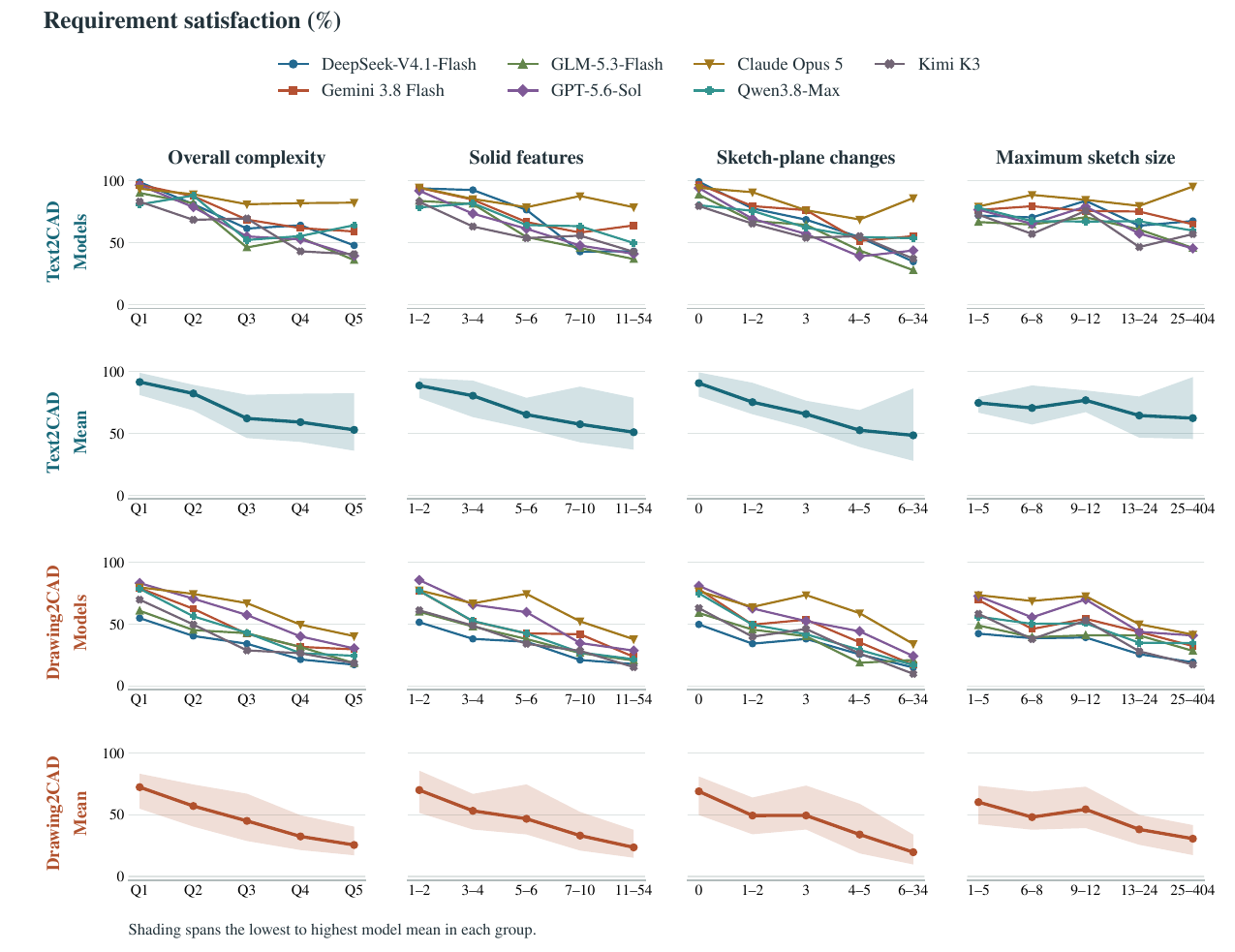}
\caption{Requirement satisfaction by construction complexity and by
individual factor. For each modality, the upper row shows individual
models and the lower row shows their mean. Shading spans the lowest to
highest model score in each group.}
\label{fig:supp-factor-trends}
\end{figure}
\endgroup

%% file: sec/supplementary/17_reproducibility.tex
\section{Experimental Configuration}
\label{app:reproducibility}
\begin{multicols}{2}
\subsection{Reasoning Effort and Build Budgets}
All seven models in the single-response comparison in Table~3 of the
main paper used high reasoning effort. Each received the task inputs
and PySW documentation and returned one program. For the agent comparison
in Table~5 of the main paper, models used low reasoning effort per call
and could revise their programs over multiple interactions.
DeepSeek Harness (v0.1.6-alpha.1) and Codex CLI (v0.154.0) allowed up to
three builds per task, including failed attempts. PySWBench used low
reasoning effort for both requirement and process evaluation.

\subsection{Token Limits and Retries}
Context windows and output limits varied by model and API. For single-response
generation, we accessed Gemini 3.8 Flash, GPT-5.6-Sol, and Claude Opus~5
through OpenRouter, using the maximum output lengths it listed at the time:
65,536, 128,000, and 128,000 tokens, respectively. Qwen3.8-Max used Alibaba
Cloud's DashScope API with a 131,072-token output limit. Standard requests
specified up to 128,000 output tokens for Kimi K3 through the Moonshot AI
API and 150,000 for DeepSeek-V4.1-Flash and GLM-5.3-Flash through the
DeepSeek and Volcengine Ark APIs, respectively. Reasoning tokens counted
toward these limits.
Across the Text2CAD and Drawing2CAD tasks, only Gemini 3.8 Flash reached
the output limit partway through writing a program. This affected 9 of its
300 outputs (3.0\%): 6 of 150 in Text2CAD and 3 of 150 in Drawing2CAD.

When a network failure interrupted a request, we retried with the same
inputs and settings until we received a response. If generation reached
the context limit or the output
was truncated, the attempt ended.
\end{multicols}